\documentclass[preprint,review,12pt]{elsarticle}
\usepackage{amssymb, amsmath, amsthm, bm}
\usepackage[noend]{algpseudocode}
\usepackage{algorithmicx,algorithm}
\usepackage{graphicx}
\usepackage{caption}
\usepackage{epstopdf}
\usepackage[colorlinks,linkcolor=blue]{hyperref}
\usepackage{bbding}
\usepackage{booktabs}
\usepackage{tikz,xcolor}
\usepackage{hyperref}
\usepackage{diagbox}

\usepackage[numbers]{natbib}
\usepackage{framed,multirow}
\usepackage{latexsym}
\usepackage{caption}
\usepackage{url}
\usepackage{bm}
\usepackage{subcaption,graphicx}
\journal{Journal}

\begin{document}
\begin{frontmatter}
\title{More Consistent Accuracy PINN via \\ Alternating Easy-Hard Training}
\author[label1] {Zhaoqian Gao \fnref{cor1}}
\author[label1] {Min Yang \corref{cor2}}
\fntext[cor1] {zqgaoalex@yeah.net}
\cortext[cor2] {Corresponding author: yang@ytu.edu.cn}
\address[label1]{School of Mathematics and Information Sciences, Yantai University, Yantai, China}

\begin{abstract}
  Physics-informed neural networks (PINNs) have recently emerged as a prominent paradigm for solving partial differential equations (PDEs), yet their training strategies remain underexplored. While hard prioritization methods inspired by finite element methods are widely adopted, recent research suggests that easy prioritization can also be effective. Nevertheless, we find that both approaches exhibit notable trade-offs and inconsistent performance across PDE types. To address this issue, we develop a hybrid strategy that combines the strengths of hard and easy prioritization through an alternating training algorithm. On PDEs with steep gradients, nonlinearity, and high dimensionality, the proposed method achieves consistently high accuracy, with relative $L^2$ errors mostly in the range of $\mathcal{O}(10^{-5})$ to $\mathcal{O}(10^{-6})$, significantly surpassing baseline methods. Moreover, it offers greater reliability across diverse problems, whereas compared approaches often suffer from variable accuracy depending on the PDE. This work provides new insights into designing hybrid training strategies to enhance the performance and robustness of PINNs.
\end{abstract}	
	
\begin{keyword}
  Physics-informed neural networks; Easy-hard prioritization; Hybrid training strategy; Alternating scheme; Consistent accuracy
\end{keyword}

\end{frontmatter}

\section{Introduction}
\label{introduction}

The rapid development of deep learning has revolutionized scientific computing, offering novel solutions to longstanding challenges in modeling complex physical systems \cite{Dong2021, Hu2024, Li2025, LiYin2024, Zhu2023}. Among these advances, physics-informed neural networks (PINNs), seamlessly integrating partial differential equations into neural network training through residual-based loss functions, have emerged as an vital framework \cite{He2024, Karniadakis2021, Lau2024, Yu2022, Zeng2024}. By bridging data-driven flexibility with physical consistency, PINNs circumvent the need for computationally expensive mesh generation and demonstrate remarkable capabilities in solving parametrized and high-dimensional partial differential equations (PDEs) \cite{Fang2022, HuKawaguchi2024, Nabian2021, Tang2023, Zhang2025}. However, their effectiveness is frequently compromised by a critical challenge: the heterogeneous contributions of loss components, which lead to unstable solution accuracy \cite{Chang2024, Gladstone2025, Jung2025, WangTeng2020, WangXu2024}.

To improve the accuracy and stability of PINNs, two seemingly distinct training strategies have been extensively studied. The first strategy, \textit{hard prioritization}, identifies and emphasizes high-loss sample points during PINN training through resampling or adaptive weighting techniques \cite{Gu2021, Liu2021, LuoYang2025, McClenny2023, Wu2023}. This strategy forces the model to focus on the more challenging regions of the PDE domain, thereby helping it capture more essential physical characteristics. For instance, residual-based adaptive sampling automatically adds more sample points in computational regions with larger residual loss \cite{Wu2023}. Luo et al. \cite{LuoYang2025} introduce Residual-based Smote (RSmote), an innovative local adaptive sampling technique tailored to improve the performance of PINNs through imbalanced learning strategies. Gu et al. \cite{Gu2021} proposed SelectNet, a self-paced learning framework that emphasizes higher-loss sample points during training. Liu and Wang \cite{Liu2021} developed a Physics-Constrained Neural Network with the Mini-Max architecture (PCNN-MM) to simultaneously update network weights (via gradient descent) and loss weights (via gradient ascent), targeting a saddle point in the weight space. McClenny and Braga-Neto \cite{McClenny2023} advanced this approach by proposing Self-Adaptive PINN (SAPINN), which adaptively assigns heavier weights to larger individual sample losses.

The second strategy, \textit{easy prioritization}, draws inspiration from human curriculum learning \cite{Bengio2009, Kong2021, WangChen2022} by initially focusing on simpler samples or tasks and gradually increasing the level of difficulty \cite{Krishnapriyan2021, Li2024, Monaco2023, WangXu2024}. This approach can reduce training instability, especially in the early stages, and promote faster convergence. Krishnapriyan et al. \cite{Krishnapriyan2021} applied curriculum learning approach in PINNs by progressively introducing high-frequency components or refining the spatiotemporal domains. Monaco and Apiletti \cite{Monaco2023} proposed a new curriculum regularization strategy to enable smooth transitions in parameter values as task difficulty increases. Wang et al. \cite{WangXu2024} addressed convection-dominated diffusion equations by assigning reduced weights to challenging regions such as boundary or interior layers to improve solution accuracy. More recently, Li et al. \cite{Li2024} introduced an Anomaly-Aware PINN (AAPINN), which enhances robustness and accuracy by identifying and excluding difficult, high-loss samples.

While both training strategies have achieved empirical success in enhancing the performance of PINNs, their relative merits and inherent trade-offs for solving PDEs remain unclear due to the lack of systematic comparative analysis. To bridge this gap, this study first conducts a comparison of these two types of training approaches based on a toy example. Our experiments reveal that the easy-prioritization method tends to emphasize the global structure of the solution -- capturing smooth, low-frequency components that dominate the overall behavior of the PDE across the entire domain \cite{Xu2020}. In contrast, hard-prioritization strategies are more inclined to focus on localized regions with higher complexity, such as areas with sharp gradients, singularities, or rapidly varying solution features, which are often more difficult to learn. Another important observation is that there is \textit{no} consistent winner between the two strategies. In some cases, the hard sample prioritization method yields better results, whereas in others, the easy sample prioritization strategy is more effective. Notably, even for the same PDE, variations in the equation coefficients can shift the advantage between the two strategies. These findings suggest that real-world PDEs, which are often complex, may not be well-suited to a one-size-fits-all training strategy.

To address this challenge, we propose a hybrid training framework that combines the strengths of both easy and hard sample prioritization. Specifically, the proposed method alternates between two training phases, each guided by a distinct optimization objective. The first phase adopts a hard prioritization strategy, employing a min-max framework to optimize a weighted PINN loss function. The second phase switches to an easy prioritization strategy, using an  anomaly-aware mechanism to progressively focus hard samples and minimize a standard (non-weighted) PINN loss. In each training epoch, these two phases are executed alternately. The alternating structure ensures a dynamic balance between easy and hard sample learning, thereby improving the robustness and generalization ability of PINNs. Given its alternating nature between easy and hard phases, we name the method Alternating Easy-Hard PINN (AEH-PINN). Experimental results demonstrate that the proposed AEH-PINN consistently surpasses existing training strategies by overcoming their accuracy trade-offs, achieving superior relative $L^2$ errors on the order of $\mathcal{O}(10^{-5})$ to $\mathcal{O}(10^{-6})$ across most challenging PDEs. The code is provided at \url{https://github.com/Gao-ST/PINN-Alternating-Easy-Hard} to facilitate reproducibility and comparison.

The remainder of this paper is organized as follows. Section 2 provides a brief overview of PINN and introduces a toy example to compare hard and easy  prioritization strategies, which motivates our proposed approach. Section 3 presents the proposed hybrid training method. Section 4 reports the experimental results on various PDEs and compares the performance of our approach with several benchmark PINN methods. Section 5 concludes the paper.

\section{Preliminaries}
\subsection{Physics-informed neural network}
	
We begin with a brief overview of PINN \cite{Raissi2019}. Consider a general PDE with initial and boundary conditions:
	\begin{equation}
		\begin{aligned}
			&\mathcal{N}[u(\bm{x},t)] = 0, (\bm{x}, t) \in \Omega\times (0,T], \\
			&\mathcal{B}[u(\bm{x},t)] = 0, (\bm{x}, t) \in \partial \Omega\times (0,T], \\
			&u(\bm{x},0) = h(\bm{x}), \bm{x}\in \Omega,
		\end{aligned}
	\end{equation}
where $u(\bm{x},t)$ denotes the solution of the PDE, $\mathcal{N}[\cdot]$ is a (possibly nonlinear) differential operator, and $\mathcal{B}[\cdot]$ represents a boundary operator that can encode various types of boundary conditions, including Dirichlet, Neumann, Robin, and periodic conditions.

The core idea of PINN is to utilize a neural network $u_{\bm{\theta}}(\bm{x},t)$, where $\bm{\theta}$ denotes trainable parameters of the network,  to approximate the solutions of PDEs $u(\bm{x},t)$.  The neural networks' output $u_{\bm{\theta}}$ provides a continuous representation that can be evaluated at any point $(\bm{x},t)$. The network is trained by minimizing the following objective function:
	\begin{equation}
		\mathcal{L}(\mathcal{X};\bm{\theta}) = \mathcal{L}_R(\bm{\theta}) + \mathcal{L}_I(\bm{\theta}) + \mathcal{L}_B(\bm{\theta}).
		\label{pinnloss}
	\end{equation}
where $ \mathcal{X}:=\{(\bm{x}_i,t_i)\}_{i=1}^{N} $ ($N=N_R+N_I+N_B $) is the training data set, and
	\begin{equation}
		\begin{aligned}
			&\mathcal{L}_R(\bm{\theta}) = \dfrac{1}{N_{R}}\sum_{i=1}^{N_R}|\mathcal{N}[u_{\bm{\theta}}(\bm{x}_i,t_i)]|^2,\quad (\bm{x}_i,t_i) \in \Omega \times  (0,T],
			\\
			&\mathcal{L}_{I}(\bm{\theta}) = \dfrac{1}{N_{I}}\sum_{i=1}^{N_I}|u_{\bm{\theta}}(\bm{x}_i ,0) - h(\bm{x}_i)|^2,\quad \bm{x}_i \in \Omega,
			\\
			&\mathcal{L}_{B}(\bm{\theta}) = \dfrac{1}{N_{B}}\sum_{i=1}^{N_B}|\mathcal{B}[u_{\bm{\theta}}(\bm{x}_i,t_i)]|^2,\quad  (\bm{x}_i,t_i) \in \partial \Omega \times  (0,T].
		\end{aligned}
	\end{equation}
These loss terms enforce the neural network to satisfy the governing PDE as well as the initial and boundary conditions.

The solution procedure for \eqref{pinnloss} is generally based on stochastic gradient descent (SGD) algorithms \cite{Robbins1951}. Owing to the non-convex nature of the optimization landscape, training is prone to convergence to local minima, which can limit the achievable accuracy. Furthermore, the loss function incorporates contributions from multiple sample points, whose loss magnitudes may vary significantly. Although the objective is to minimize the average loss, individual sample points can still exhibit large residuals even when the overall loss is small. This imbalance undermines the reliability of the learned solution, particularly in multiscale scenarios.

As outlined in Section \ref{introduction}, two distinct strategies have been developed to mitigate this problem: (1) hard prioritization and (2) easy prioritization. In the following subsection, we are to introduce a toy example to conduct a comparative analysis of these strategies.

\subsection{Hard or easy prioritization}
\label{part1}
\begin{figure}[H]
	\hspace{-1em}
	\includegraphics[width=\linewidth]{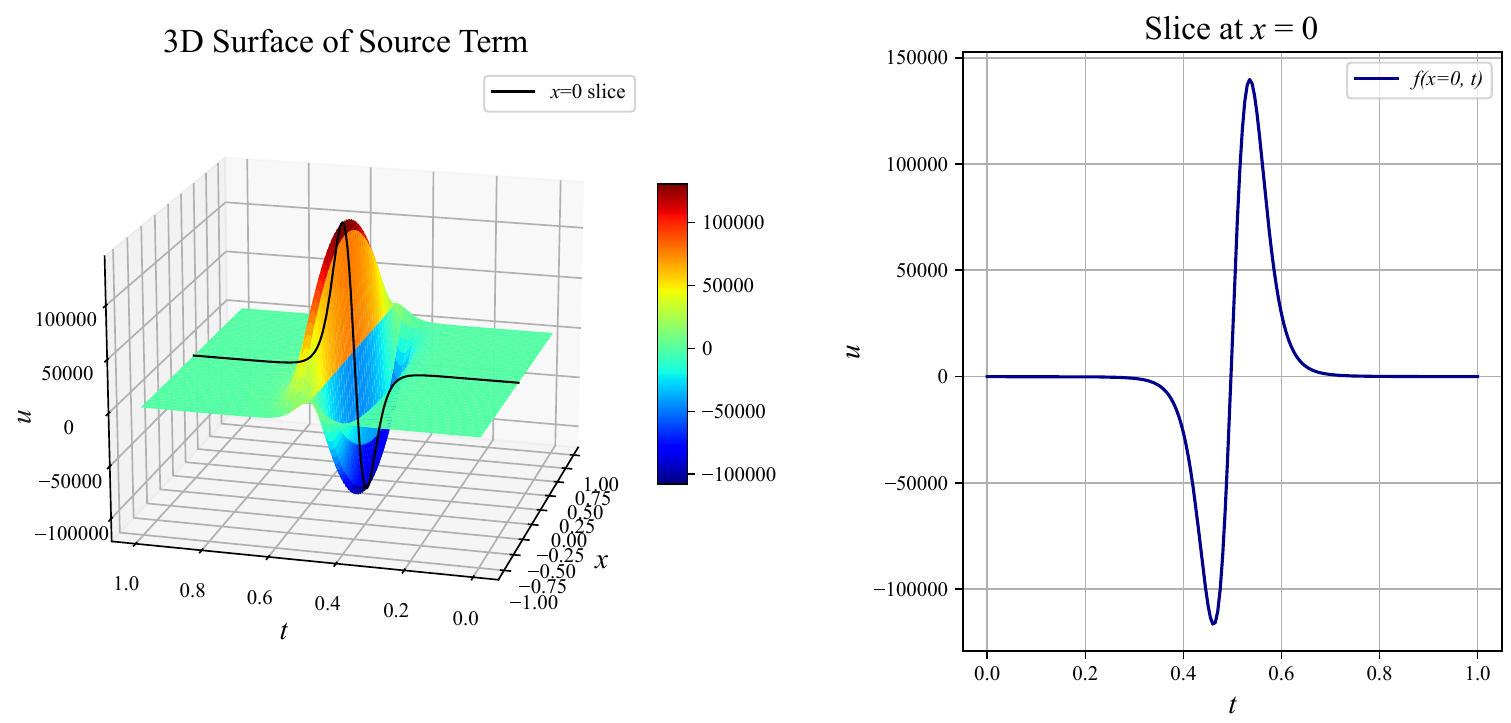}
    \caption{
		\textbf{Visualization of a source term at $ \alpha=0.11 $ used in the heat conduction equation}.
		\textbf{Left:} 3D surface of the source term $f(x,t)$, exhibiting sharp localized peaks and steep gradients, with value ranges exceeding $10^5$.
		\textbf{Right:} 1D slice of $f(x,t)$ along $x = 0$, showing highly nontrivial temporal behavior.
		Such source terms introduce strong local features and multiscale variations in the solution, posing significant challenges for standard PINNs to learn effectively.
	}
	\label{ST}
\end{figure}

Consider the following heat conduction problem with steep gradients, adapted from \cite{WangYao2024}:
\begin{equation}
\label{toy}
\begin{aligned}
			&u_t = u_{xx} + f(x,t), &x \in (-1,1),\; t \in (0,1],
			\\
			&u(x,0) = (1-x^2)e^{\frac{1}{1+\alpha}}, &x \in (-1,1),
			\\
			&u(-1,t) = u(1,t) = 0, &t \in (0,1],
\end{aligned}
\end{equation}
where the exact solution is  set as $u(x,t) = (1-x^2) e^{\frac{1}{(2t-1)^2+\alpha}}$. The source term $f(x,t)$, visualized in Figure~\ref{ST}, exhibits sharp localized peaks and steep gradients. As $\alpha$ decreases, the solution becomes increasingly sharp, presenting a challenging test case for numerical methods.

For comparison, we consider two baseline methods. The first is SAPINN \cite{McClenny2023}, a hard-sample-prioritization method that introduces an adversarial min-max optimization framework. The overall objective is formulated as:
\begin{equation}
	\mathcal{L}(\mathcal{X};\bm{\theta},\bm{w}) \triangleq  \mathcal{L}_R(\bm{\theta},\bm{w}_R)+\mathcal{L}_I(\bm{\theta},\bm{w}_I)+\mathcal{L}_B(\bm{\theta},\bm{w}_B),
	\label{r1}
\end{equation}
where the three components correspond to the residual loss, initial condition loss and boundary condition loss, respectively:
	\begin{equation}
		\begin{aligned}
			\mathcal{L}_R(\bm{\theta},\bm{w}_R) &= \frac{1}{2} \sum_{i=1}^{N_R} m(w_i) \left| \mathcal{N}[u_{\bm{\theta}}(\bm{x}_i,t_i)] \right|^2,
			\\
			\mathcal{L}_I(\bm{\theta},\bm{w}_I) &= \frac{1}{2} \sum_{i=1}^{N_I} m(w_i) \left| u_{\bm{\theta}}(\bm{x}_i,0) - h(\bm{x}_i) \right|^2,
			\\
			\mathcal{L}_B(\bm{\theta},\bm{w}_B) &= \frac{1}{2} \sum_{i=1}^{N_B} m(w_i) \left| \mathcal{B}[u_{\bm{\theta}}(\bm{x}_i,t_i)] \right|^2.
		\end{aligned}
		\label{r2}
	\end{equation}
Here $ \bm{w}_R, \bm{w}_I, \bm{w}_B  $ are trainable weights, and $m( \cdot )$ is a non-negative self-adaptation mask function defined on $[0,\infty)$. SAPINN simultaneously updates the network parameters $\bm{\theta} $  via gradient descent and the sample weights $ \bm{w} $  via gradient ascent:
	\begin{equation*}
		\bm{\theta} \leftarrow \bm{\theta}  - \eta  \nabla_{\bm{\theta} } \mathcal{L}(\mathcal{X} ;\bm{\theta}, \bm{w}), \quad
		\bm{w} \leftarrow \bm{w} + \rho  \nabla_{\bm{w}} \mathcal{L}(\mathcal{X};\bm{\theta}, \bm{w}).
	\end{equation*}
This adversarial learning scheme adaptively reallocates attention toward regions with higher residuals and aims to improve  the network's ability to capture sharp transitions in the solution.

The second method is AAPINN \cite{Li2024}, which introduces a dynamic anomaly exclusion mechanism during training. It builds upon the standard PINN loss function \eqref{pinnloss} and computes the mean and standard deviation of the sample-wise feedforward losses:
\begin{align}
\label{statistic}
	\Psi(\bm{\theta}) = \frac{1}{N} \sum_{i=1}^{N} \mathcal{L}_i(\bm{\theta}), \quad	\sigma^2(\bm{\theta}) = \frac{1}{N-1} \sum_{i=1}^{N} \left( \mathcal{L}_i(\bm{\theta}) - \Psi(\bm{\theta}) \right)^2
\end{align}
where $\mathcal{L}_i(\bm{\theta})$ is the feedforward loss of the $i$-th sample.
Based on these statistics, AAPINN periodically detects outliers with exceptionally high losses. If no such samples are identified, the model is trained on the full dataset. Otherwise, the training focuses only on a subset of easy and moderate samples, excluding the identified anomalies.

\begin{figure}[hbtp]
	\centering
	\includegraphics[width=0.7\linewidth]{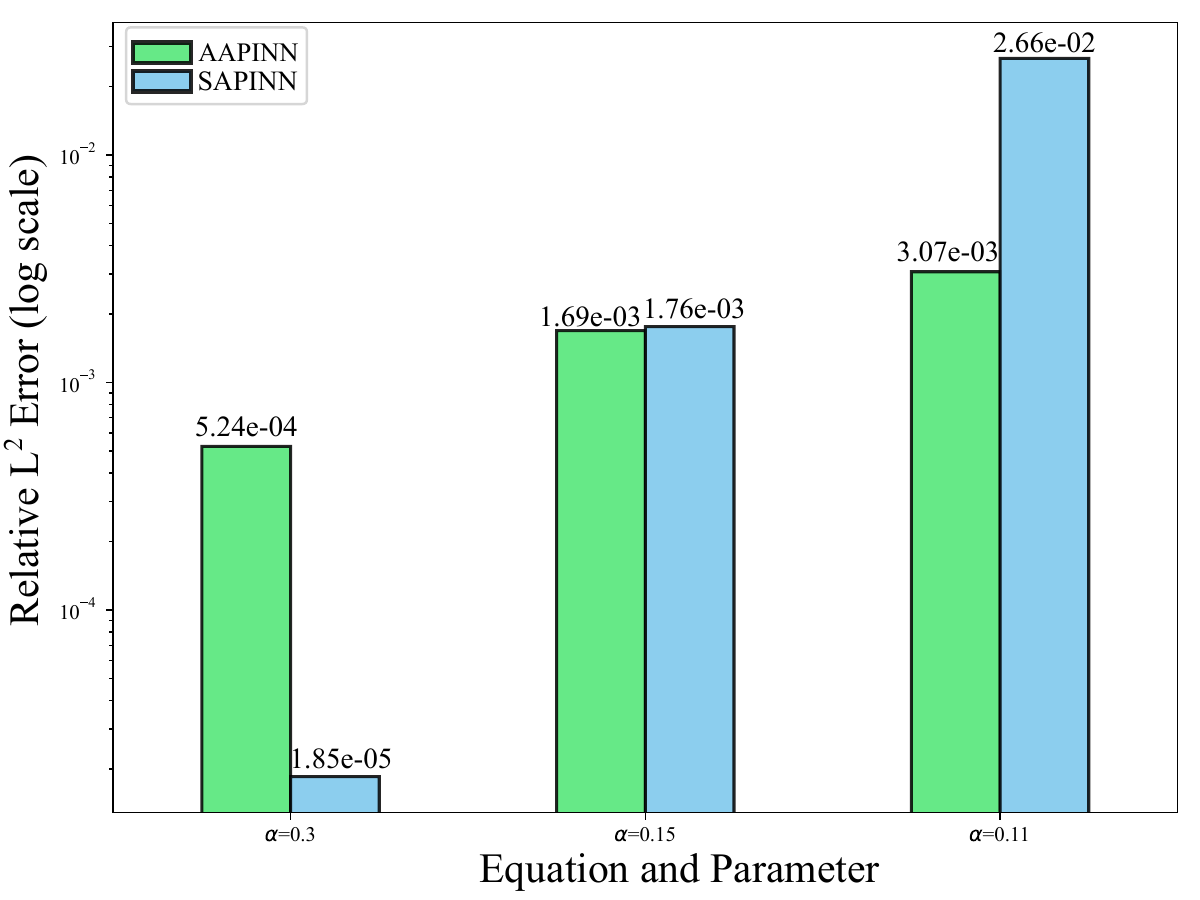}
	\caption{Relative $L^2$ errors of SAPINN and AAPINN on equation \eqref{toy} with different difficulty levels ($\alpha = 0.3, 0.15, 0.11$).}
	\label{fig:screenshot002}
\end{figure}

We apply SAPINN and AAPINN to solve the equation \eqref{r1} under varying levels of difficulty, controlled by the parameter $\alpha = 0.3, 0.15, 0.11$. Both models employ the same architecture: a fully connected neural network with four hidden layers, each containing 50 neurons and activated by the Tanh function. Training is performed using the Adam optimizer \cite{Kingma2015}.

As shown in Figure~\ref{fig:screenshot002}, neither method consistently outperforms the other across all difficulty levels. For moderate difficulties ($\alpha = 0.3$), SAPINN achieves notably higher accuracy, reaching a relative $L^2$ error of $1.85\times10^{-5}$ compared to AAPINN's $5.24\times10^{-4}$, a reduction by a factor of approximately 28.3. However, this advantage reverses at $\alpha = 0.11$, where AAPINN attains a relative error of $3.07\times10^{-3}$, outperforming SAPINN's $2.66\times10^{-2}$ by a factor of 8.7. These results suggest that, even for the same type of PDE, neither hard  prioritization nor easy prioritization strategy can maintain universally superior performance.

\begin{figure}[htbp]
	\centering
	\includegraphics[width=\linewidth]{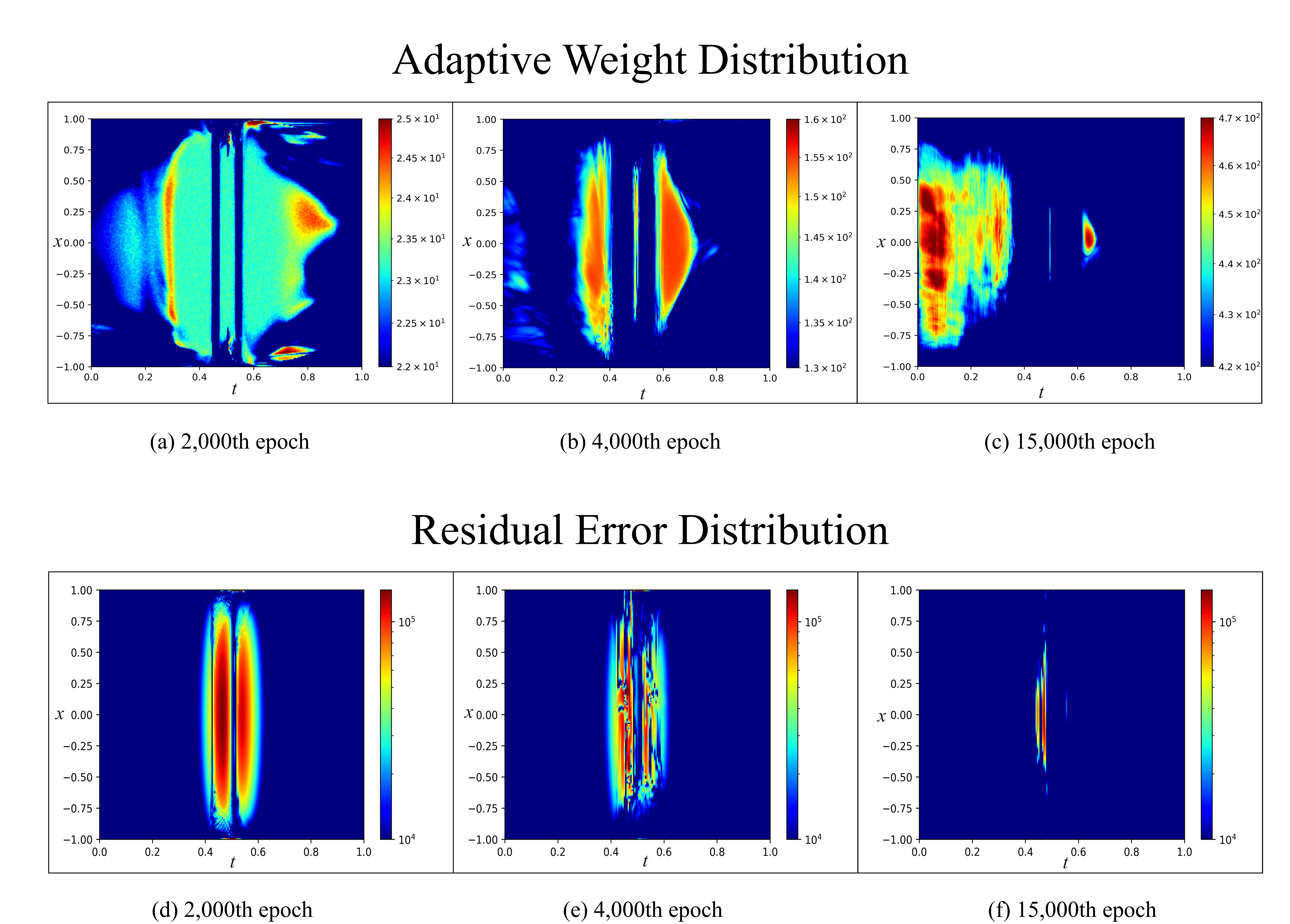}
	\caption{Dynamics of weight and error distribution in SAPINN for equation  ($\alpha=0.11$) at different training stages.}
	\label{fig:combinsed}
\end{figure}

We note that as two fundamentally different learning mechanisms, AAPINN mitigates the impact of extreme outliers by eliminating highly difficult samples and focusing on easier ones. Over time, it gradually shifts attention toward more challenging regions. This strategy avoids overemphasis on problematic samples and promotes training stability, especially during the early stages (as will be further illustrated in the training curves in Section \ref{losscurve}). In contrast, SAPINN emphasizes learning from difficult samples. As shown in Figure~\ref{fig:combinsed}, the relative $L^2 $ error and weight distributions at different training stages reveal that SAPINN assigns significantly larger weights to a small subset of samples with high residuals. While this boosts the approximation accuracy in those specific regions, the overall prediction performance remains suboptimal: some low-weighted regions experience increased approximation errors.

Given that computationally difficult regions are usually limited in proportion, hard prioritization strategies tend to act as ``local" methods. In contrast, easy-sample prioritization can be viewed as a more ``global" approach that better accounts for general sample behavior. However, both strategies involve inherent trade-offs: hard prioritization may hinder convergence due to overemphasis on difficult samples, whereas easy prioritization may fail to adequately resolve complex regions of the solution space. We argue that a hybrid of these two approaches may yield more consistent and robust performance for PINNs. In the next section, we will introduce our proposed method based on this idea.

\section{The Proposed Method}
\label{AEHPINN}

In order to balance the inherent trade-offs between easy and hard prioritization strategies in training PINNs, we propose a hybrid framework that integrates both paradigms. This framework consists of two key phases.
The first is a weighted adversarial phase, inspired by SAPINN \cite{McClenny2023}, which prioritizes difficult regions by dynamically adjusting sample weights through a min-max optimization process. The second is an easy-prioritization phase, drawing from the ideas in AAPINN \cite{Li2024}, which excludes extreme difficult samples to stabilize training.

To effectively combine these two phases, we design an alternating training scheme that iteratively switches between the hard- and easy-prioritization stages. Rather than simply connecting the two phases in a fixed sequential or parallel fashion, this alternating strategy promotes continual interaction between local refinement and global stability throughout the training process.

While the specific implementation is novel in the context of PINNs, the underlying intuition echoes broader insights from curriculum learning and self-paced learning in computer vision and natural language processing \cite{Graves2017, Jiang2015}. Jiang et al. \cite{Jiang2015}  discover the missing link between CL and SPL, and propose a unified framework named self-paced curriculum leaning, which takes into account both prior knowledge known before training and the learning progress during training. Graves et al. \cite{Graves2017} introduce a method for automatically selecting the path, or syllabus, that a neural network follows through a curriculum so as to maximise learning efficiency. These studies suggest that dynamically balancing sample difficulty over time, rather than committing to a single difficulty regime, can lead to more robust and generalizable models. To the best of our knowledge, this is the first work to introduce such a structured alternating training mechanism into PINNs.

Specifically, in the first phase, we emphasize difficult samples by employing a min-max optimization scheme. Penalty weights are assigned to each training sample $\{(\bm{x}_i,t_i)\}_{i=1}^{N}$ and iteratively updated to focus on high-loss regions. The objective is formulated as:

\begin{equation}
	\label{HardPhase}
	\begin{aligned}
		\underset{\bm{\theta}}{\min}~\underset{\bm{w}}{\max}~\mathcal{\hat{L}}(\mathcal{X};\bm{\theta},\bm{w})
		\triangleq  &\frac{1}{N_R} \sum_{i=1}^{N_R} w_{i} \left| \mathcal{N}[u_{\bm{\theta}}(\bm{x}_i,t_i)] \right|^2\\
		+ &\frac{1}{N_{I}} \sum_{i=1}^{N_{I}} w_{i} \left| u_{\bm{\theta}}(\bm{x}_i,0) - h(\bm{x}_i) \right|^2\\
		+ &\frac{1}{N_{B}} \sum_{i=1}^{N_{B}} w_{i}  \left| \mathcal{B}[u_{\bm{\theta}}(\bm{x}_i,t_i)] \right|^2,
	\end{aligned}
\end{equation}
where $w_{i}$ denotes the sample-wise weights. Unlike the original SAPINN formulation~\eqref{r2}, our implementation omits the use of the mask function, resulting in a simplified objective.

The penalty weight updates are performed via gradient ascent:
\begin{equation}
	\label{weightup}
	\bm{w} \leftarrow \bm{w} + \eta  \nabla_{\bm{w}} \mathcal{\hat{L}}(\mathcal{X};\bm{\theta}, \bm{w}).
\end{equation}
where $\eta$ is the learning rate for the weight update. This encourages the network to place more emphasis on samples associated with larger residuals. The network parameters and weights are jointly updated in parallel using the gradients of the weighted loss.:
\begin{equation}
	\label{netup}
	\bm{\theta} \leftarrow \bm{\theta} - \eta  \nabla_{\bm{\theta}} \mathcal{\hat{L}}(\mathcal{X}; \bm{\theta}, \bm{w}).
\end{equation}

In the second phase, the lack of batch training in SGD poses a challenge, often leading to unstable updates and poor model generalization. Inspired by \cite{Li2024}, we address this issue by dynamically prioritizing easy samples through a easy-to-hard sampling strategy. Besides, instead of using the weighted loss function as in \eqref{HardPhase}, we adopt the standard PINN loss \eqref{pinnloss} without any sample weighting. 

Different from the sample selection mechanism in \eqref{statistic}, we do not rely on statistical heuristics to select samples. Instead, we sort all samples based on their feedforward loss values and retain only the top $r$ fraction with the lowest losses. These selected samples form a subset $\mathcal{X}_{sub}$, which is used for updating the network parameters via:
\begin{equation}
     \bm{\theta} \leftarrow \bm{\theta}- \eta \nabla_{\bm{\theta}} \mathcal{L}(\mathcal{X}_{sub}; \bm{\theta})
\end{equation}

The selection ratio $r$ is determined as follows. Let $\texttt{cycle}$ denote the current position within a training cycle, defined as the current epoch index modulo the update period $P$. The sample selection ratio $r$ is then dynamically adjusted according to:
\begin{equation}
       r = \min\{0.5 + 0.99 \cdot \frac{\texttt{cycle}}{P},\ 1\}, \quad \text{where }\texttt{cycle} = 1, 2, \ldots, P.
\end{equation}
Clearly, the value of $r$ increases progressively from $50\%$ to $100\%$, allowing more samples to participate in training as the cycle progresses. By using the period $P$ to reset $r$ periodically, the model restarts with 50\% of the easy samples every $P$ epochs and gradually increases the sample ratio. This way may explore the solution space more randomly and helps avoid potential local minima.

Morevoer, it is worth noting that a key difference between our second phase and AAPINN \cite{Li2024} lies in the treatment of difficult samples. AAPINN operates under an outlier assumption and permanently excludes a small subset of extremely difficult samples throughout training, which may result in information loss. In contrast, our approach ensures that all training samples are eventually incorporated as $r$ reaches 1. This full-sample utilization mitigates the risk of discarding potentially informative sample points.

The entire training procedure alternates between two phases in a repeated manner: Phase 1 runs for $S_1$ steps, followed by Phase 2 for $S_2$ steps. This alternating cycle continues until convergence or until the maximum number of epochs is reached. The complete pseudo-code is presented in Algorithm~\ref{alg:hybrid}.

\begin{algorithm}[H]
	\caption{Pseudo-code of the proposed AEH-PINN}
	\label{alg:hybrid}
	\begin{algorithmic}[1]
		\State \textbf{Initialization:} Network parameters $\bm{\theta}_0$, sample weight $w_i = 1\ \forall i$, learning rate $\eta$, phase steps $S_1$, $S_2$, \texttt{cycle} period $P$, maximum training steps \texttt{max\_epochs}.
		\While{\texttt{epoch} $\leq$ \texttt{max\_epochs}}
		
		\State \textbf{Phase 1.} Objective: $\underset{\bm{\theta}}{\min}~\underset{\bm{w}}{\max}~\mathcal{\hat{L}}(\mathcal{X}_{full};\bm{\theta},\bm{w})$
		\For{$s = 1$ to $S_1$}
			\State $\bm{w} \leftarrow \bm{w}+ \eta \nabla_{\bm{w}} \mathcal{\hat{L}}(\mathcal{X}_{full};\bm{\theta},\bm{w})$
			\State $\bm{\theta} \leftarrow \bm{\theta} - \eta \nabla_{\bm{\theta}} \mathcal{\hat{L}}(\mathcal{X}_{full};\bm{\theta},\bm{w})$
		\EndFor
		
		\State \textbf{Phase 2.}	Objective: $\underset{\bm{\theta}}{\min}~\mathcal{L}(\mathcal{X}_{sub};\bm{\theta})$
		\State $\texttt{cycle} \leftarrow \texttt{epoch}\% P$
		\For{$s = 1$ to $S_2$}
			\State $r \leftarrow \min\left\{0.5 + 0.99 \cdot \dfrac{\texttt{cycle}}{P},\ 1 \right\}$
			\State $\mathcal{X}_{sub} \leftarrow$ top-$r\%$ of samples sorted by $\mathcal{L}_i(\bm{\theta})$ (ascending)
			\State $\bm{\theta} \leftarrow \bm{\theta} - \eta \nabla_{\bm{\theta}} \mathcal{L}(\mathcal{X}_{sub};\bm{\theta})$
		\EndFor
		
		\State \texttt{epoch} $\leftarrow$ \texttt{epoch} + 1
		\EndWhile
	\end{algorithmic}
\end{algorithm}

\section{Experiments}
	
\label{experiments}
\subsection{Setup}
We evaluate the proposed method in this section. The PINN architecture consists of four hidden layers with 50 neurons each, initialized using He initialization \cite{He2015}. The Tanh activation function is used throughout the network. For comparison, we consider several PINN baselines:

\begin{enumerate}
    \item \textbf{RAD} \cite{Wu2023}: Dynamically allocates more sampling points to regions with higher residual losses, enabling the model to focus on areas where it performs poorly.

    \item \textbf{SelectNet} \cite{Gu2021}: A learning framework that emphasizes harder samples by adaptively adjusting sample weights. It updates weights and network parameters in an alternating manner.

    \item \textbf{CL-Reg} \cite{Monaco2023}: Decomposes the training of a complex PDE into a sequence of related PDEs with incrementally increasing difficulty. Unlike other compared methods, it is a task-based approach rather than a sample-based PINN training method.

    \item \textbf{SAPINN} \cite{McClenny2023}: Assigns adaptive weights to each sample point and updates both weights and network parameters simultaneously, in contrast to SelectNet's alternating scheme.

    \item \textbf{AAPINN} \cite{Li2024}: Introduces anomaly-aware progressive learning by dynamically filtering out high-loss samples during training, allowing the model to focus on more reliable sample points.
\end{enumerate}

We evaluate our method on three challenging linear PDEs, each presenting distinct numerical difficulties: (1) a heat conduction equation with steep gradients, (2) the Helmholtz equation exhibiting oscillatory behavior, and (3) a singularly perturbed convection-diffusion equation. Beyond these, we consider two nonlinear PDEs known for their complexity: (1) the Allen-Cahn equation and (2) the Sine-Gordon equation. To further test scalability and robustness, we also apply our method to a four-dimensional multiscale PDE that combines high dimensionality with complex multiscale features.

The configuration of sampling points for each problem is summarized in Table~\ref{diff}.

\begin{table}[!htb]
    \centering
    \caption{The configuration of sampling points for each problem.}
    \renewcommand{\arraystretch}{1}
        \begin{tabular}{lccc}
            \toprule
            	Problem								& \#Residual points & \#Boundary points & \#Test points \\
            \midrule
	            Heat (steep gradients)			& 54,756 & 3,600 & 40,401
	            \\
	            Helmholtz						& 54,756 & 3,600 & 20,301
	            \\
	            Convection-dominated diffusion  & 2,501  & 2     & 1,001
	            \\
	            Allen-Cahn  & 12,800  & 3,600     & 20,301
	            \\
	            Sine-Gordon  & 12,800  & 4,800    & 10,000
	            \\
	            4D multiscale  & 12,800  & 2,400     & 10,000
	            \\
            \bottomrule
        \end{tabular}
    \label{diff}
\end{table}

For our method, we set the number of inner-loop iterations to $S_1 = 10$ and $S_2 = 1$ for Phase 1 and Phase 2, respectively. In Phase 2, the sample selection ratio is updated with a cycle period of $P = 300$.  The network is trained using the Adam optimizer \cite{Kingma2015} in float32 precision with a initial learning rate of $10^{-3}$.  All baseline methods are implemented with their original default hyperparameters unless otherwise specified. The maximum number of training epochs is fixed at 100,000 to ensure convergence for all tested PDEs, especially those with sharp gradients or multiscale behavior.

We evaluate all methods using the relative $L^2$ error:
\begin{equation}
    \text{ReL2} := \frac{\sqrt{\sum_{i=1}^{N} |u_{\bm{\theta}}(\bm{x}_i) - u(\bm{x}_i)|^2}}{\sqrt{\sum_{i=1}^{N}|u(\bm{x}_i)|^2}}
\end{equation}
where $u_{\bm{\theta}}(\bm{x})$ denotes the predicted solution, $u(\bm{x})$ represents the ground truth, and $N$ is the number of uniformly sampled test points. To ensure statistical reliability, each experiment is repeated 10 times, and the average performance is reported.

\subsection{Heat conduction with steep gradients}
\label{heat_section}
Consider the following heat conduction problem with steep gradients \cite{WangYao2024}:
\begin{equation}
    \label{Heat}
		\begin{aligned}
			&u_t = u_{xx} + f(x,t), 				 &x \in (-1,1),\; t \in (0,1],
			\\
			&u(x,0) = (1-x^2)e^{\frac{1}{1+\alpha}}, &x \in (-1,1),
			\\
			&u(-1,t) = u(1,t) = 0, 					 &t \in (0,1],
		\end{aligned}
\end{equation}
where $u(x,t)$ is the unknown solution, $f(x,t)$ is a given source term, and Dirichlet boundary conditions are imposed at the spatial boundaries.

Following \cite{WangYao2024}, the exact solution is defined as $u(x,t) = (1 - x^2) e^{\frac{1}{(2t - 1)^2 + \alpha}}$ with $\alpha = 0.11$. This solution exhibits extremely steep gradients near the interior of the domain $\Omega$, where the function values rise sharply from near zero to over 8000. Such behavior creates a highly stiff and localized profile, presenting a serious challenge for both traditional numerical solvers and PINN-based methods.

\begin{figure}[hbt]
	\centering
	\includegraphics[width=1\textwidth]{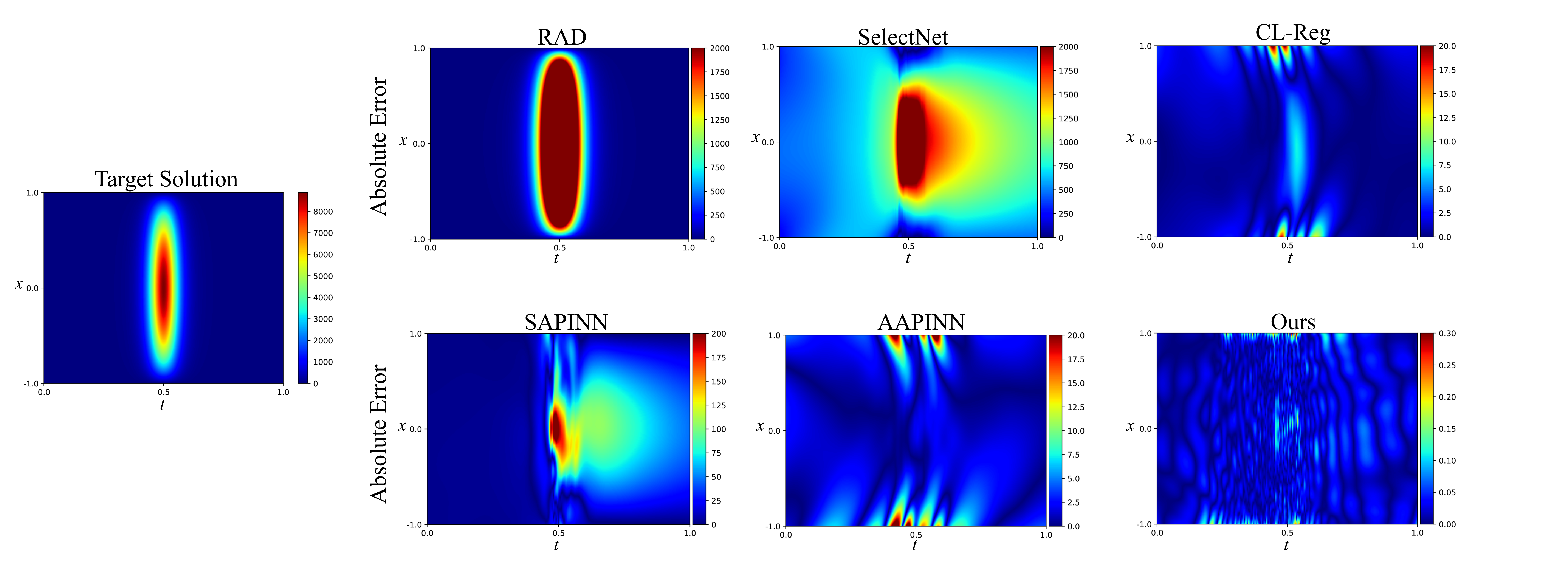}
	\caption{Absolute error distribution comparison of different methods for the heat conduction problem \eqref{Heat}.}
	\label{result_heat}
\end{figure}

Figure~\ref{result_heat} demonstrates that hard prioritization strategies, such as RAD, SelectNet, and SAPINN, tend to yield relatively poor accuracy on this problem. In contrast, easy prioritization approaches like CL-Reg and AAPINN exhibit significantly improved performance, benefiting from their progressive or anomaly-aware learning schemes.

\begin{table}[htb]
	\centering
	\caption{Relative $L^2$ errors of various methods for the heat conduction problem \eqref{Heat}}
	\resizebox*{\linewidth}{!}{
		\begin{tabular}{lcccccc}
			\toprule
			Method &RAD \cite{Wu2023} &SelectNet \cite{Gu2021} &CL-Reg \cite{Monaco2023} &SAPINN \cite{McClenny2023} &AAPINN \cite{Li2024} & Ours
			\\
			\midrule
			ReL2   &9.94e-01  &5.79e-01  &\underline{9.17e-04}  &2.66e-02  &2.59e-03 &\textbf{1.05e-5}
			\\
			\bottomrule
		\end{tabular}
	}
	\label{heat}
\end{table}

Table~\ref{heat} quantitatively compares the relative $L^2$ errors achieved by different methods. hard prioritization methods generally produce errors on the order of $10^{-1}$ to $10^{-2}$, indicating difficulty in resolving this steep gradients problem. In comparison, CL-Reg and AAPINN achieve substantially lower errors of $9.17 \times 10^{-4}$ and $2.59 \times 10^{-3}$, respectively. Notably, the proposed AEH-PINN outperforms all baselines, achieving a relative $L^2$ error of $1.05 \times 10^{-5}$ and successfully capturing the solution's sharp internal transitions with a significant advantage.

\subsection{Helmholtz equation}
\label{helm_section}

Consider the following Helmholtz equation, which models wave propagation and acoustics:
\begin{equation}
\label{Helm}
		\begin{aligned}
			-&\Delta u - k^2 u = f,  &&\bm{x}\in\Omega= (0,1)^2,
			\\
			&u(\bm{x}) = 0,      &&\bm{x}\in\partial \Omega,
		\end{aligned}
\end{equation}
where $u$ denotes the wave field, $k$ is the wave number controlling the oscillatory nature of the solution, and $f$ is a known source term. The problem is defined on the unit square $\Omega = (0,1)^2$ with homogeneous Dirichlet boundary conditions imposed on $\partial \Omega$.

The Helmholtz equation poses significant challenges for numerical methods, especially as the wave number $k$ increases. This is primarily due to the so-called \emph{pollution effect}, where standard discretization methods require increasingly fine meshes to resolve the oscillations accurately, resulting in high computational cost. In our setup, following \cite{Lu2021}, we use the exact solution $u(x, y) = \sin(kx)\sin(ky)$ with $k = 4\pi$, which corresponds to a moderately high-frequency regime and serves as a stringent test for the accuracy and resolution capability of the proposed method.

\begin{figure}[hbt]
	\centering
	\includegraphics[width=1\textwidth]{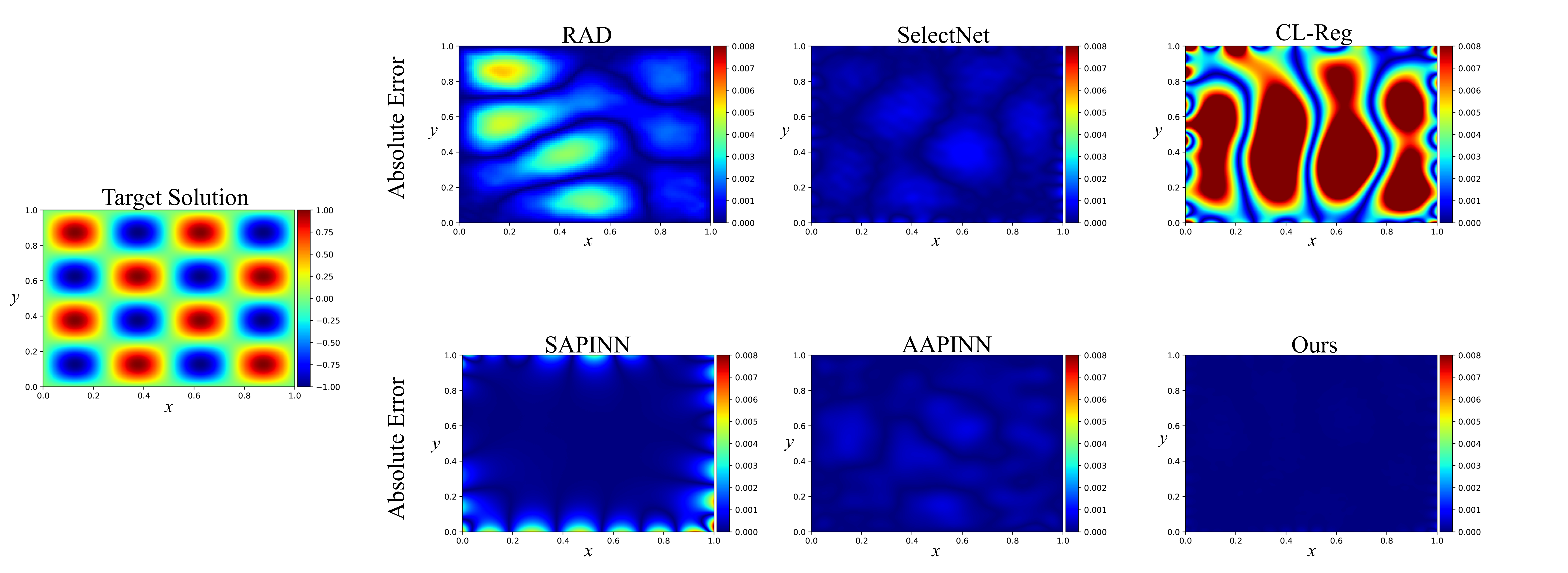}\hfill
	\caption{Absolute error distribution comparison of different methods for the Helmholtz equation \eqref{Helm}.}
	\label{helm}
\end{figure}

Compared to the heat conduction problem \eqref{Heat}, the solution of the Helmholtz equation varies within a much narrower range, between 0 and 1, making it relatively less challenging. As shown in Figure~\ref{helm}, most methods, including both hard prioritization and easy prioritization approaches, yield comparable levels of accuracy. The only notable exception is CL-Reg, which, despite being the second-best performer in the heat conduction task, exhibits substantially inferior accuracy in this case. This observation suggests that current PINN methods may exhibit problem-dependent behavior, performing well in some cases but less effectively in others.

\begin{table}[!htb]
	\centering
	\caption{Relative $L^2$ errors of various methods for the Helmholtz equation \eqref{Helm}.}
	\label{helm_rel2}
	\resizebox{\textwidth}{!}
	{
		\begin{tabular}{ccccccc}
			\toprule
			Method &RAD \cite{Wu2023} &SelectNet \cite{Gu2021} &CL-Reg \cite{Monaco2023} &SAPINN \cite{McClenny2023} &AAPINN \cite{Li2024} &Ours\\
			\midrule
			ReL2   &3.06e-03 &4.97e-04 &1.47e-02 &1.66e-03 &\underline{4.54e-04}  &\textbf{6.01e-05}\\
			\bottomrule
	\end{tabular}}
\end{table}

Furthermore, Table~\ref{helm_rel2} reveals that our proposed AEH-PINN method still achieves a clear advantage in accuracy. It attains a relative $L^2$ error of $6.01\times10^{-5}$, which is significantly better than the second-best result from AAPINN ($4.54\times10^{-4}$) and has an improvement of 7 times. This performance is also consistent with the previous experiment, where the errors  maintain an order of $10^{-5}$.

\subsection{1D Convection-dominated diffusion equation}
\label{1d_section}

Consider the following convection-diffusion problem \cite{WangXu2024}:
\begin{equation}
\label{CDD}
\begin{aligned}
    -\epsilon u_{xx} + (x-2)u_x &= f(x), \quad x \in (0,1), \\
    u(0) &= u(1) = 0,
\end{aligned}
\end{equation}
where $u$ denotes the solution and $f$ is the source term. Dirichlet boundary conditions are imposed.

This problem is a classical example of a \emph{singularly perturbed equation}, where the small diffusion parameter $\epsilon \ll 1$ leads to a \emph{sharp boundary layer} near $x = 0$. As $\epsilon$ decreases, the solution becomes increasingly steep in this region, with gradients on the order of $\mathcal{O}(1/\epsilon)$. Such behavior poses significant challenges:
\begin{itemize}
    \item Standard discretization methods (e.g., uniform grids with finite differences or finite elements) may fail to resolve the steep boundary layer unless extremely fine meshes are used near $x = 0$, resulting in high computational cost.
    \item Moreover, common PINNs tend to struggle with such layer behavior \cite{Chang2024,Krishnapriyan2021}, especially when the gradient scale is far beyond the global resolution captured by the neural network.
\end{itemize}
Thus,  the problem  is a \emph{demanding test case} for assessing the robustness of PINN-based methods. We set $\epsilon = 10^{-6}$ and take the exact solution as
\begin{equation*}
       u(x) = \cos\left(\frac{\pi x}{2}\right)\left(1 - e^{-\frac{2x}{\epsilon}}\right).
\end{equation*}

\begin{figure}[htbp]
	\centering
	\includegraphics[width=\textwidth]{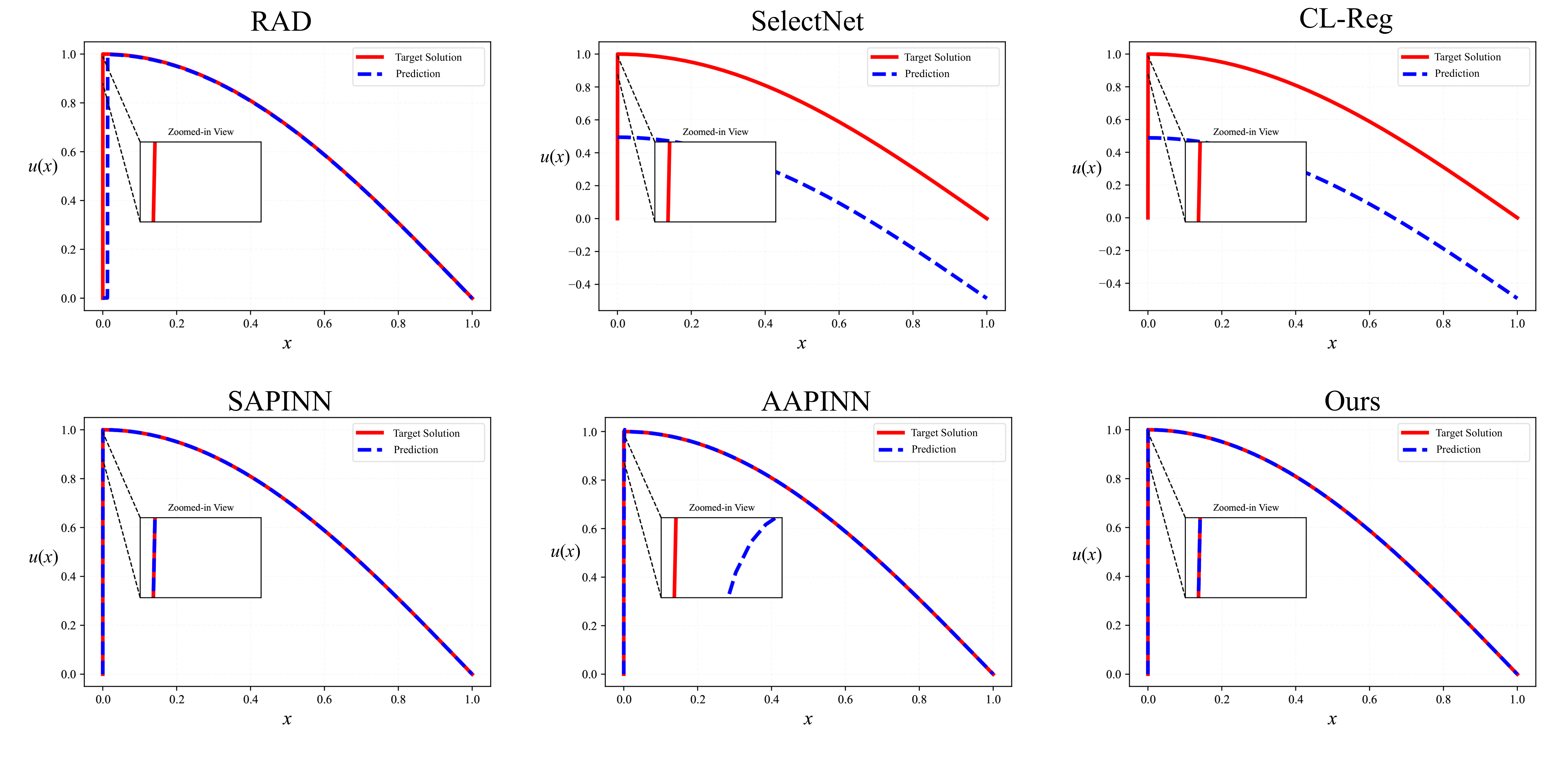}
	\caption{\textbf{Visualization of predicted solutions for the 1D convection-dominated equation~\eqref{CDD}}. The hard prioritization method outperforms the easy prioritization strategy in capturing the locally challenging region.}
	\label{fig:combined}
\end{figure}

Figure~\ref{fig:combined} provides a visual comparison of the approximated solutions produced by various methods. hard prioritization approaches such as RAD and SelectNet struggle to resolve sharp-gradient regions, while the easy prioritization method AAPINN accurately captures the global trend but lacks local precision. The task-based method CL-Reg, on the other hand, fails to deliver an adequate approximation across both global and local features.

\begin{table}[!htb]
	\centering
	\caption{Relative $L^2$ errors of various methods for the convection-dominated equation \eqref{CDD}.}
	\label{1d_rel2}
	\resizebox{\linewidth}{!}{
		\begin{tabular}{ccccccc}
			\toprule
			Method & RAD \cite{Wu2023} &SelectNet \cite{Gu2021} & CL-Reg \cite{Monaco2023}& SAPINN \cite{McClenny2023} &AAPINN \cite{Li2024}   & Ours\\
			\midrule
			ReL2   &1.60e-01 &7.12e-01 &7.14e-01 &\underline{3.90e-06} &5.16e-03 &\textbf{2.79e-06}\\
			\bottomrule
	\end{tabular}}
\end{table}

Quantitative results presented in Table~\ref{1d_rel2} show that both SAPINN and our proposed method significantly outperform the other baselines, achieving relative errors on the order of $10^{-6}$--about three orders of magnitude lower than competing methods. Among the two, our method yields slightly better accuracy, consistently attaining the lowest errors.

\subsection{Allen-Cahn equation}

To evaluate the performance of our proposed method on nonlinear PDEs, we consider the Allen-Cahn equation \cite{Lu2021}:
\begin{equation}
\label{AC}
	u_t = d \cdot u_{xx} + 5(u - u^3), \quad x \in (-1,1), \ t \in (0,1]
\end{equation}
where $ u $ is the scalar field of interest, and $ d > 0 $ is the diffusion coefficient, controlling the strength of spatial diffusion relative to the nonlinear reaction term. The initial condition is defined as:
\begin{equation*}
	u(x, 0) = x^2 \cos(\pi x)
\end{equation*}
and the Dirichlet boundary conditions are specified as:
\begin{equation*}
	u(-1, t) = u(1, t) = -1
\end{equation*}

The Allen-Cahn equation is widely studied in the PINN literature due to its pronounced nonlinear behavior, particularly the emergence of sharp internal transition layers and high sensitivity to time discretization in classical numerical schemes~\cite{Li2024,McClenny2023}. In our experiments, we set the diffusion coefficient $d = 0.001$, which leads to narrow interfaces and poses significant challenges for accurate approximation.

\begin{figure}[htbp]
	\centering
	\includegraphics[width=\linewidth]{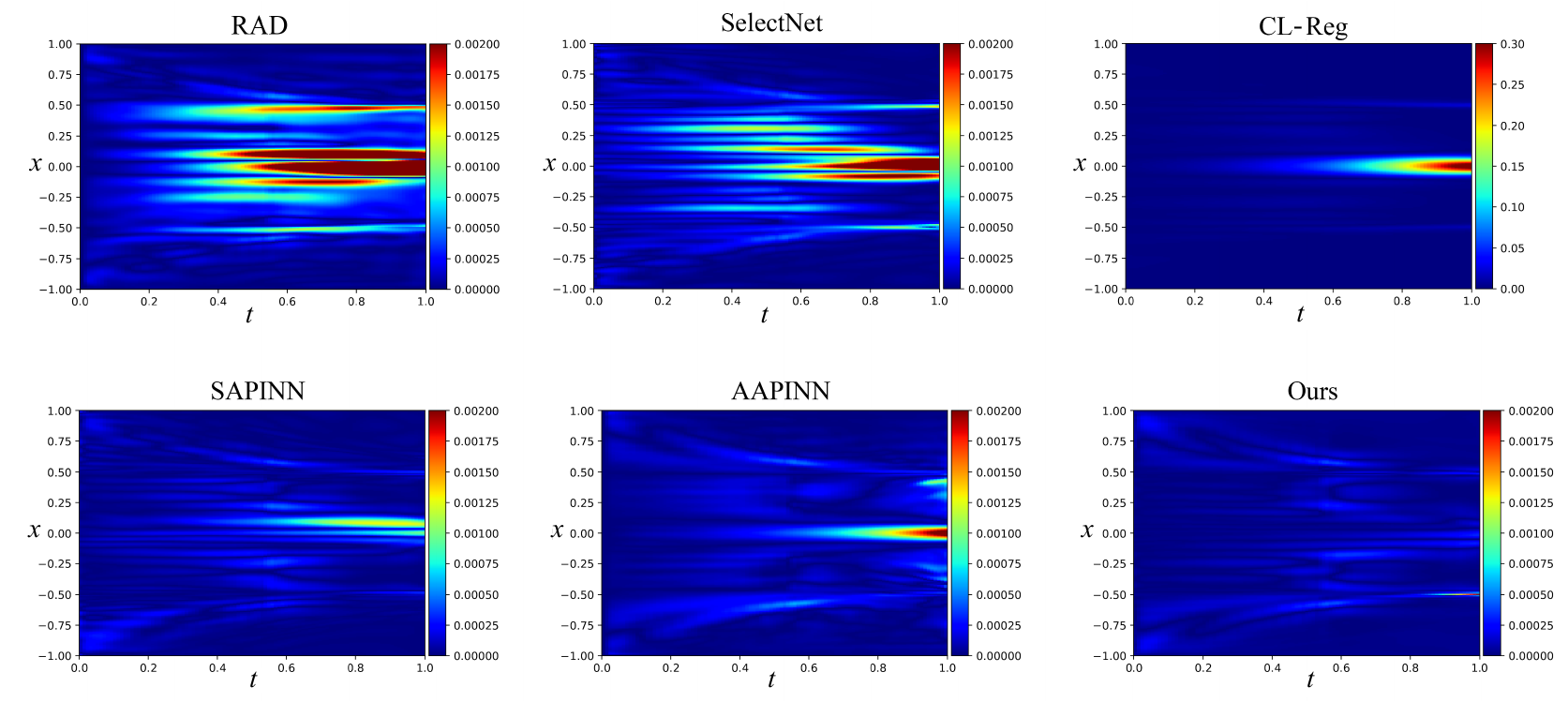}

	\caption{Absolute error distribution comparison of different methods for the Allen-Cahn equation \eqref{AC}.}
	\label{fig:ac_error}
\end{figure}

\begin{figure}[htbp]
	\centering
	\begin{minipage}{0.4\textwidth}
		\centering
		\includegraphics[width=\linewidth]{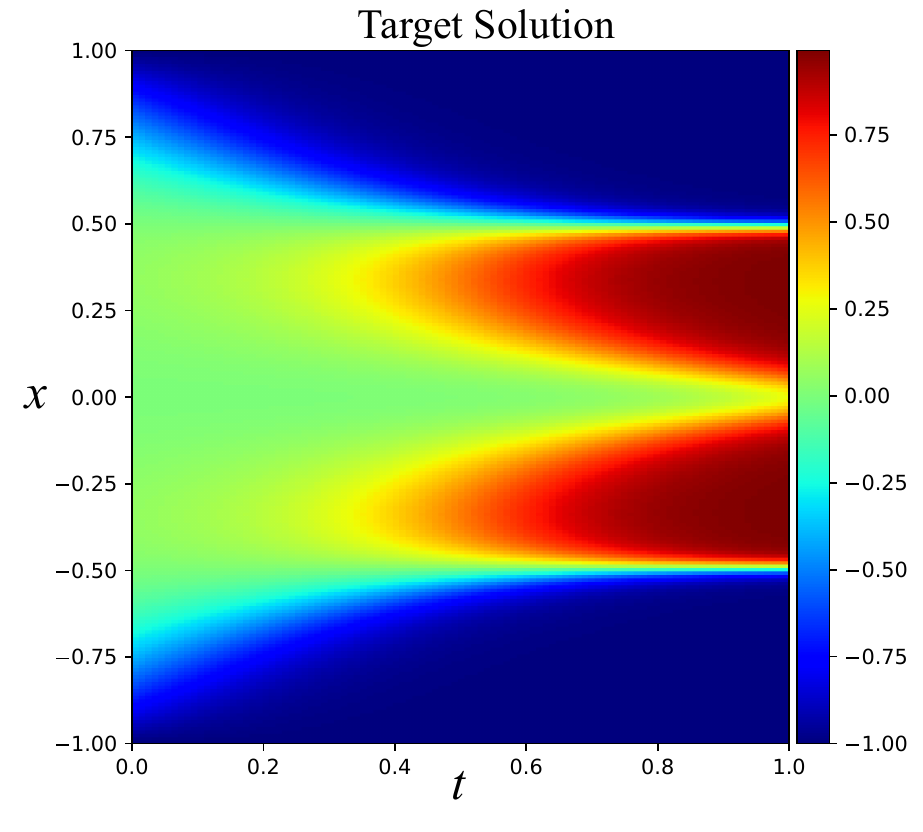}
		\captionsetup{labelformat=empty}
		\caption*{(a)}
		\label{fig:ac1}
	\end{minipage}
	\hspace{2em}
	\begin{minipage}{0.38\textwidth}
		\centering
		\includegraphics[width=\linewidth]{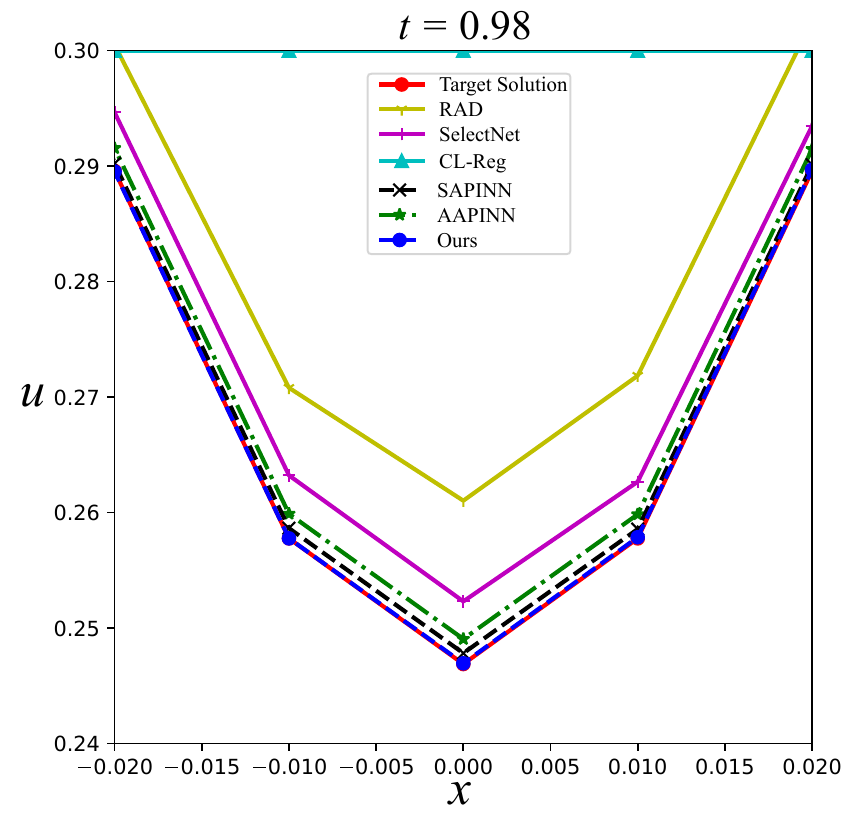}
		\captionsetup{labelformat=empty}
		\caption*{(b)}
		\label{fig:ac2}
	\end{minipage}
	\caption{\textbf{Compare of different methods conducted on Allen-Cahn equation}. (a) Target solution of Allen-Cahn equation. (b) Local region zoomed at $ t=0.98 $.}
	\label{fig:ac_zoom}
\end{figure}

Figure~\ref{fig:ac_error} illustrates the absolute errors produced by each method across the spatio-temporal domain. It highlights the regions where each method struggles and demonstrates how our approach successfully resolves critical transition zones. Figure~\ref{fig:ac_zoom} presents a zoomed-in comparison at $t = 0.98$, where our method attains the best fitting, whereas the baseline methods exhibit notable deviation.

We summarize the relative $L^2$ errors of all methods in Table~\ref{ac_rel2}. The results show that our hybrid method achieves the lowest prediction error ($8.12 \times 10^{-5}$) among all compared approaches. While RAD, AAPINN, and CL-Reg show some ability to approximate the solution, they struggle to capture the sharp interface as effectively as our model. The hard-prioritization strategy SAPINN($ 2.11\times 10^{-4} $) demonstrates superior capability in resolving the challenging regions, outperforming the easy-prioritization strategy AAPINN($ 2.76\times 10^{-4} $) in terms of accuracy.

\begin{table}[!htb]
	\centering
	\caption{Relative $L^2$ errors of various methods for the Allen-Cahn equation \eqref{AC}.}
	\label{ac_rel2}
	\resizebox{\linewidth}{!}{
		\begin{tabular}{ccccccc}
			\toprule
			Method & RAD~\cite{Wu2023} & SelectNet~\cite{Gu2021} & CL-Reg~\cite{Monaco2023} & SAPINN~\cite{McClenny2023} & AAPINN~\cite{Li2024} & Ours \\
			\midrule
			ReL2   & 1.32e-03 & 6.58e-04 & 3.87e-02 & \underline{2.11e-04} & 2.76e-04 & \textbf{8.12e-05} \\
			\bottomrule
		\end{tabular}
	}
\end{table}

\subsection{Sine-Gordon equation}
Another type of nonlinear PDE we consider is the two-dimensional Sine-Gordon equation \cite{HuKawaguchi2024}:
\begin{equation}
\label{Sine-Gordon}
	\begin{aligned}
		&\Delta u(\bm{x}) + \sin(u(\bm{x})) = g(\bm{x}), \quad \bm{x} \in (-4,4)^2,
	\end{aligned}
\end{equation}
with the boundary condition
\begin{equation}
	\begin{aligned}
		& u(-4,x_2) = -\sin(4 x_2)  \cdot(1-\cos(16+x_2^2)),
		\\
		& u(4,x_2)  =  \sin(4 x_2)  \cdot(1-\cos(16+x_2^2)),
		\\
		& u(x_1,-4) = -\sin(4x_1)  \cdot(1-\cos(16+x_1^2)),
		\\
		& u(x_1,4)  = \sin(4 x_1)   \cdot(1-\cos(16+x_1^2)).
	\end{aligned}
\end{equation}

This Sine-Gordon equation poses considerable challenges throughout most regions of the domain. The key difficulty stems from the strong nonlinearity of the $\sin(u)$ term and the tight coupling among variables, which often leads to complex, spatially varying solution structures. Traditional numerical solvers may suffer from instability or require dense meshes to maintain accuracy, while standard PINN training strategies frequently fail to converge or do so at a prohibitively slow rate.

To evaluate model robustness in such settings, we adopt the exact solution
\[
u(x_1,x_2) = \sin(x_1 \cdot x_2)\cdot(1 - \cos(x_1^2 + x_2^2)),
\]
which is highly nonlinear, non-separable, and strongly coupled across both variables. Unlike simpler expressions such as $u(x_1,x_2) = f(x_1+x_2)$, which can be reduced to a one-dimensional problem via coordinate transformation, the solution does not admit such simplification. Instead, it involves complex pairwise interactions between $x_1$ and $x_2$ that give rise to rich local structures and intricate spatial patterns. These features make the problem difficult and highlight the need for training methods that are reliable and flexible.

\begin{figure}[htbp]
	\centering
	\setlength{\tabcolsep}{0.5em}
	\renewcommand{\arraystretch}{1.2}
	\setlength\fboxrule{0.8pt}
	\setlength\fboxsep{2pt}
	\begin{tabular}{cc}
		\fbox{\includegraphics[width=0.45\textwidth, height=5.3cm]{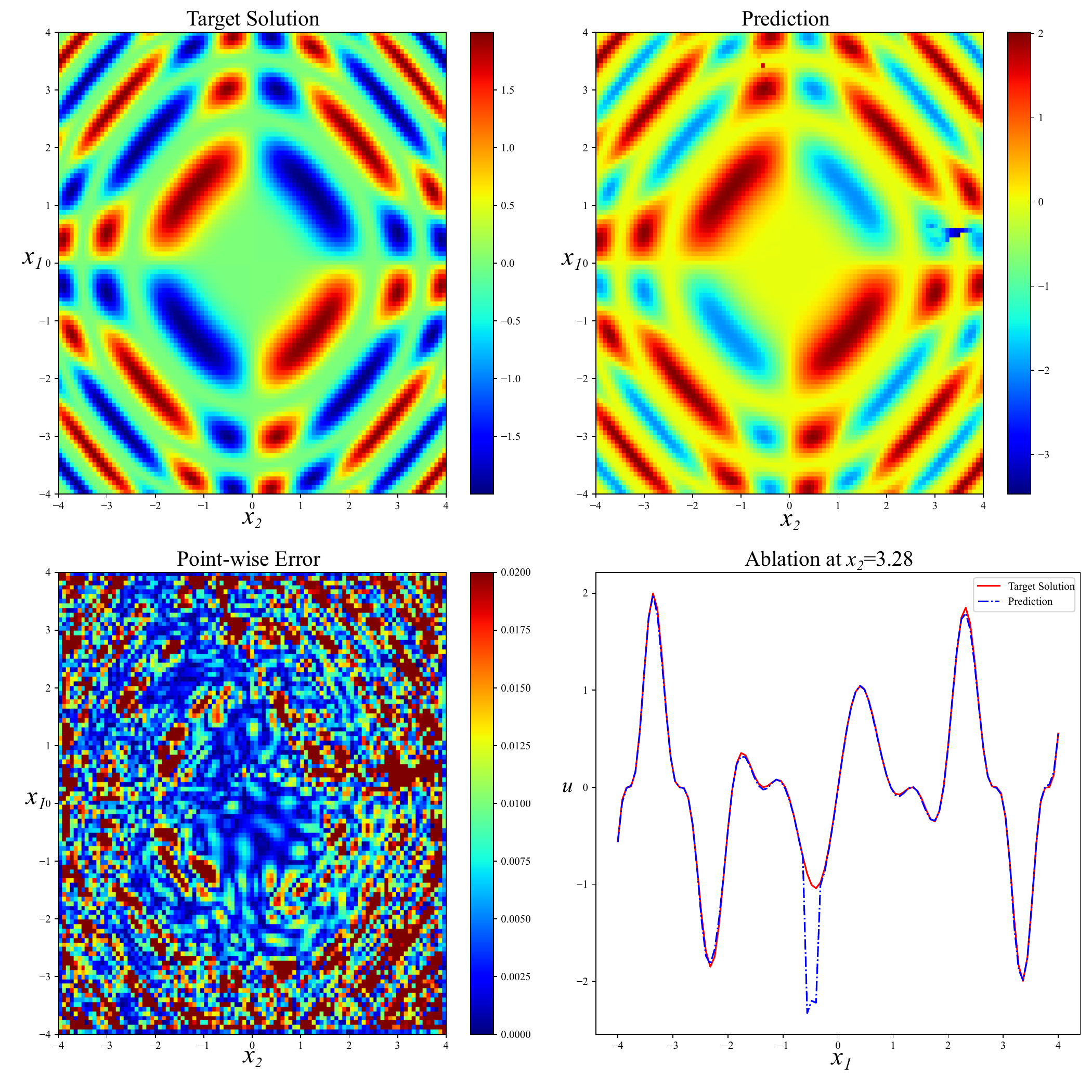}} &
		\fbox{\includegraphics[width=0.45\textwidth, height=5.3cm]{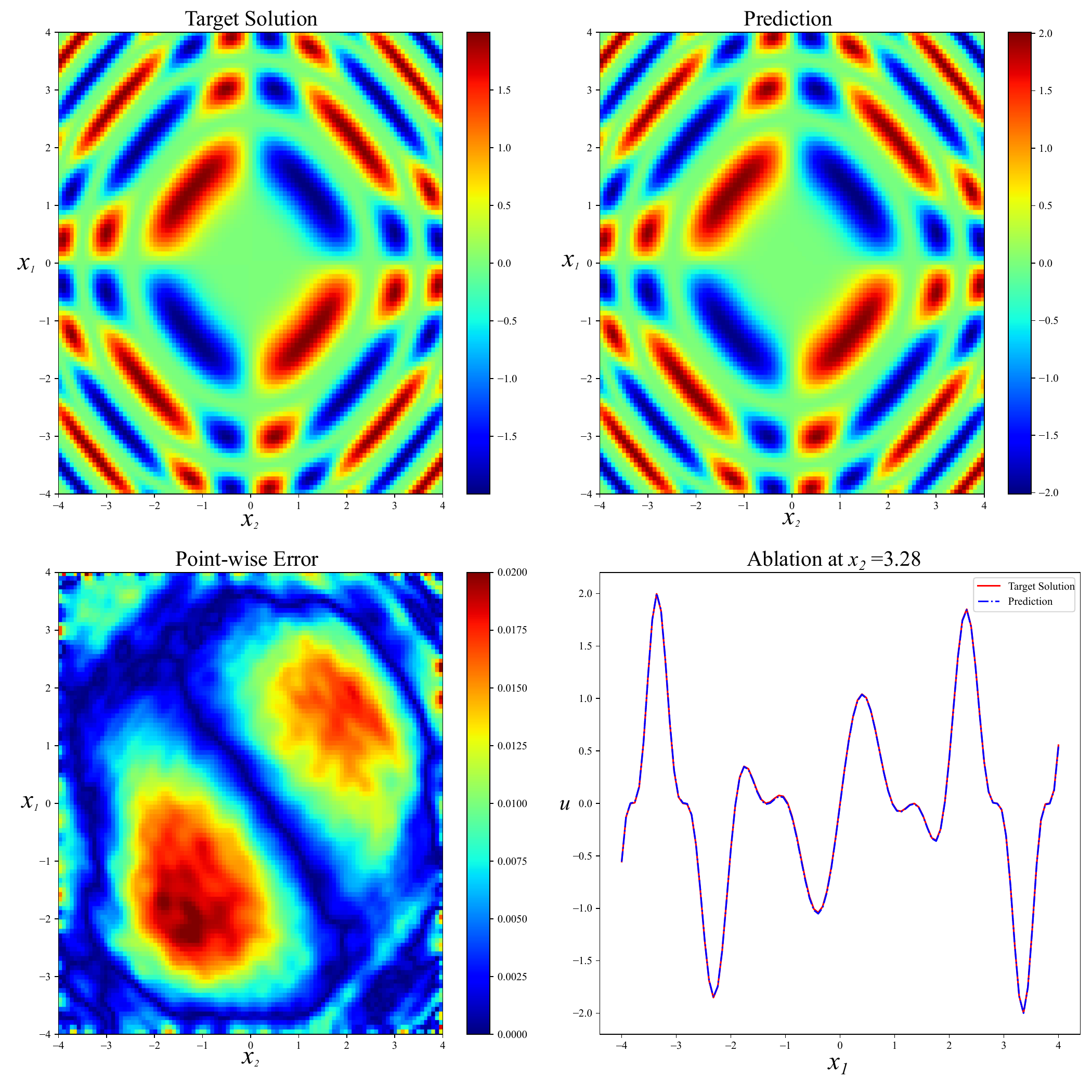}} \\
		(a) RAD & (b) SelectNet \\[0.8em]
		
		\fbox{\includegraphics[width=0.45\textwidth, height=5.3cm]{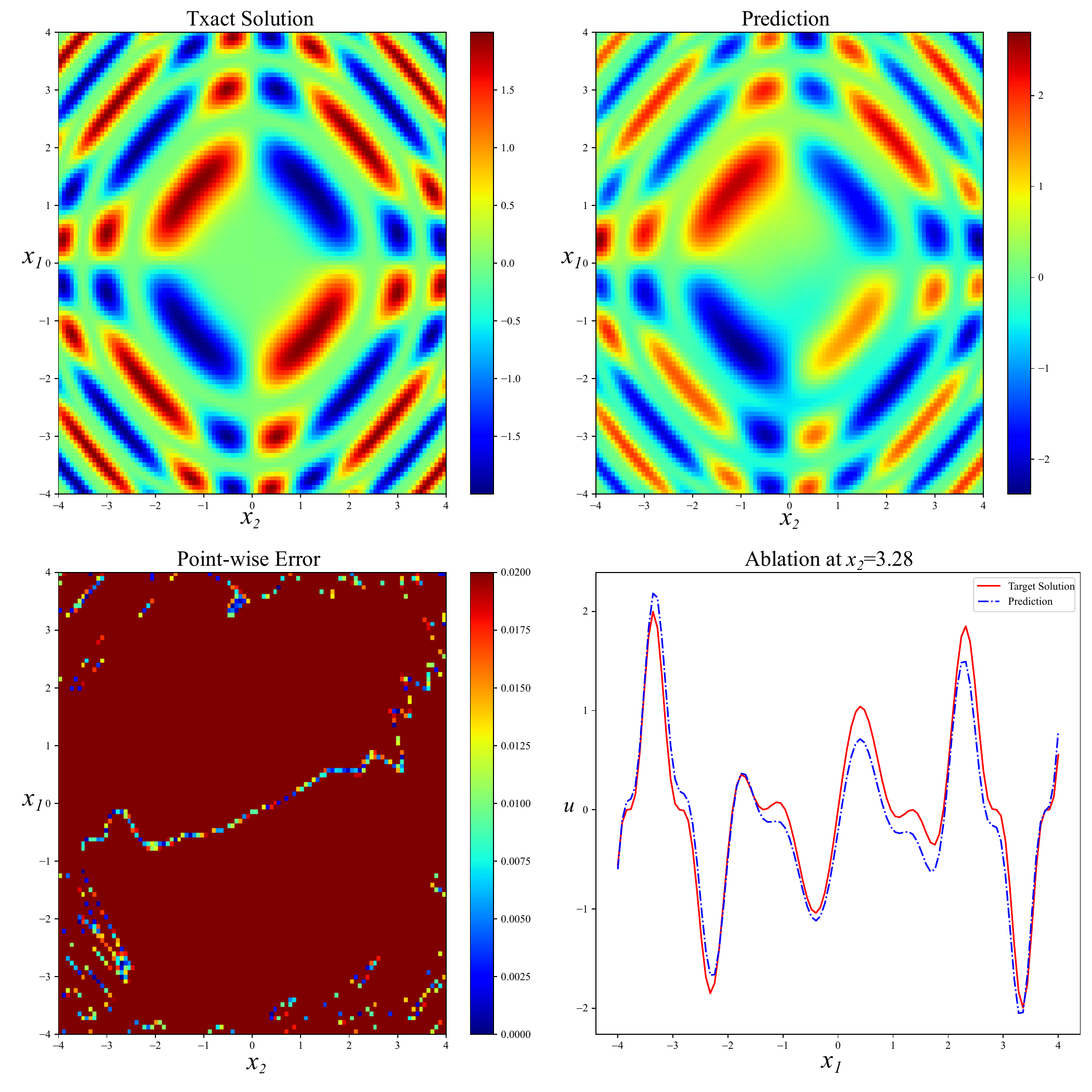}} &
		\fbox{\includegraphics[width=0.45\textwidth, height=5.3cm]{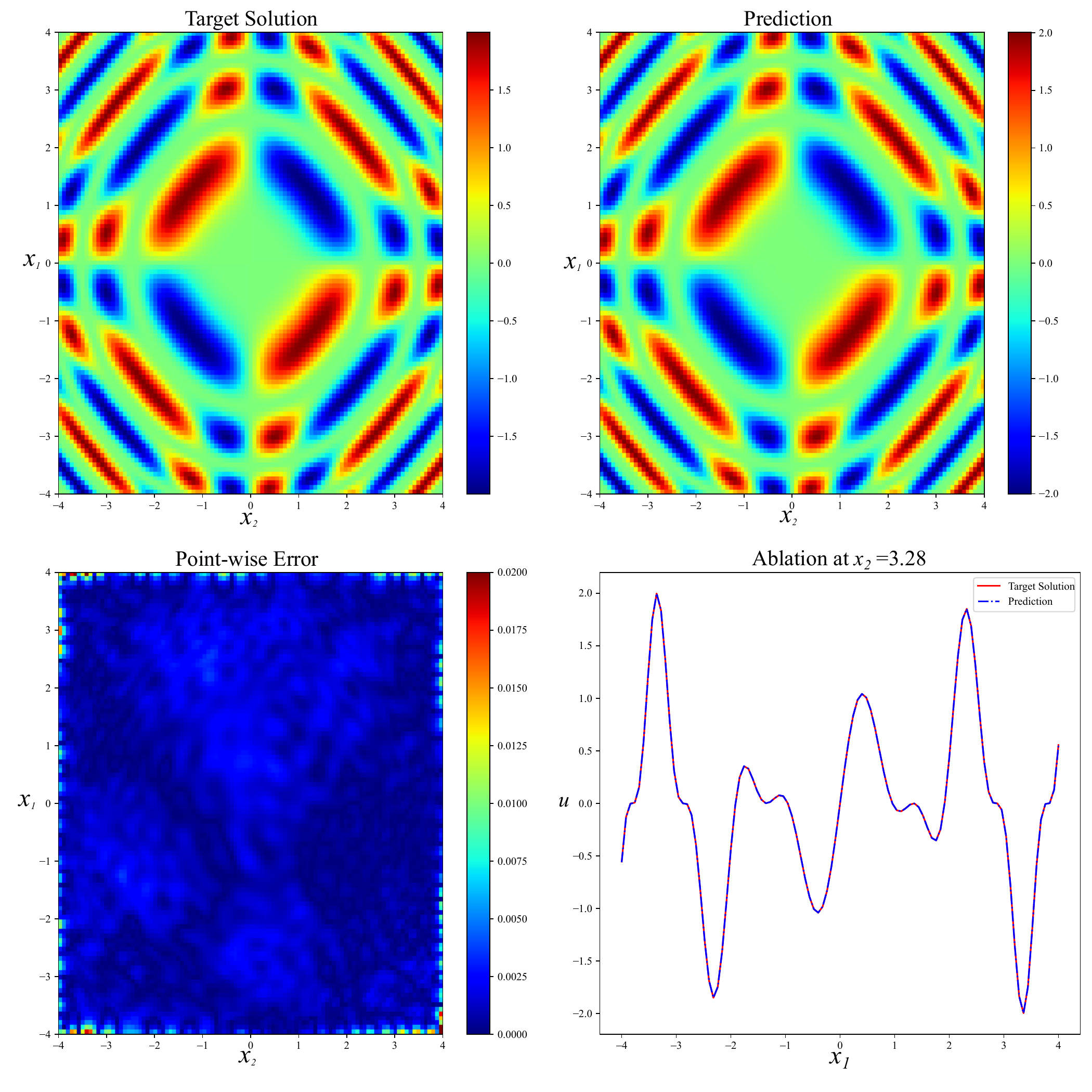}} \\
		(c) CL-Reg & (d) SAPINN \\[0.8em]
		
		\fbox{\includegraphics[width=0.45\textwidth, height=5.3cm]{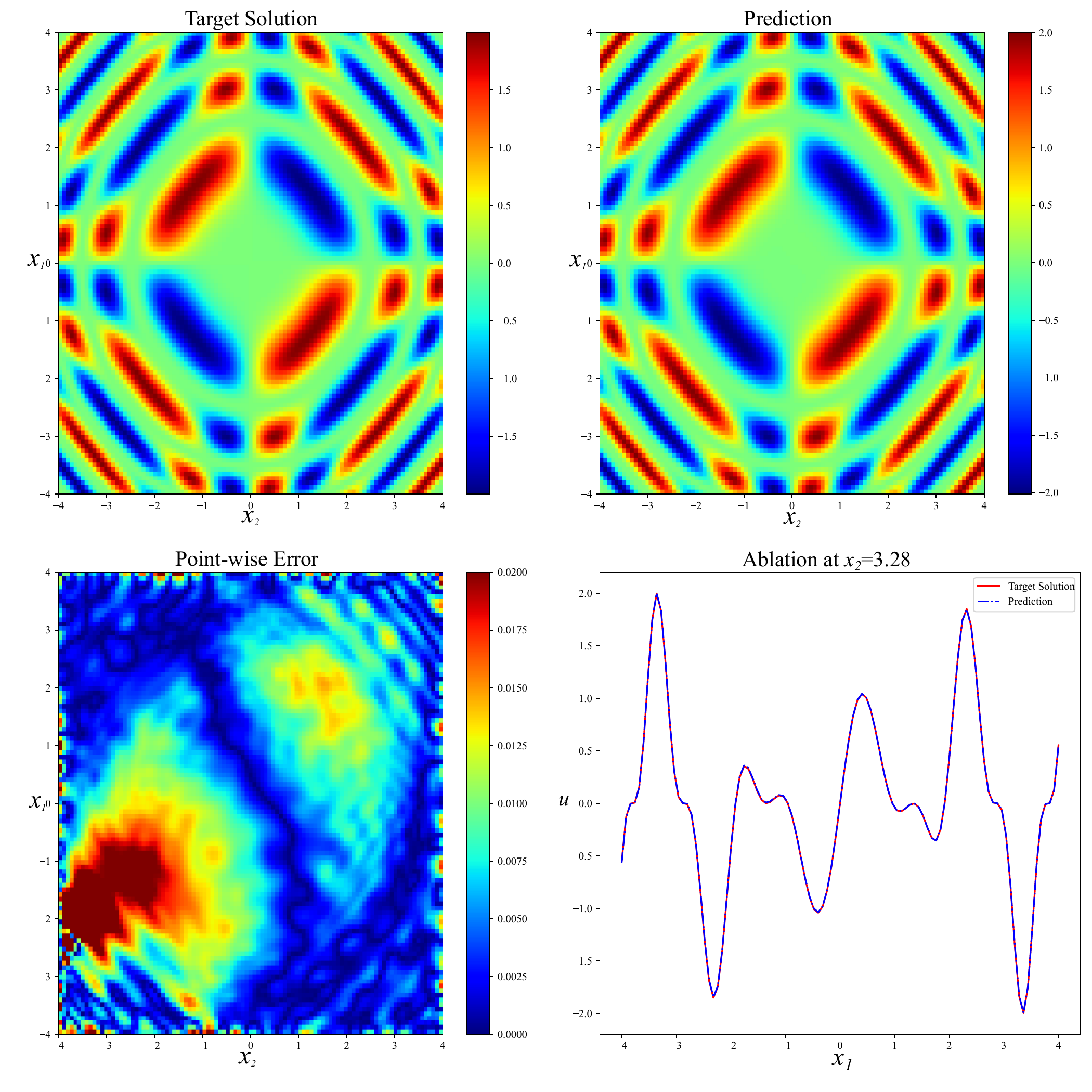}} &
		\fbox{\includegraphics[width=0.45\textwidth, height=5.3cm]{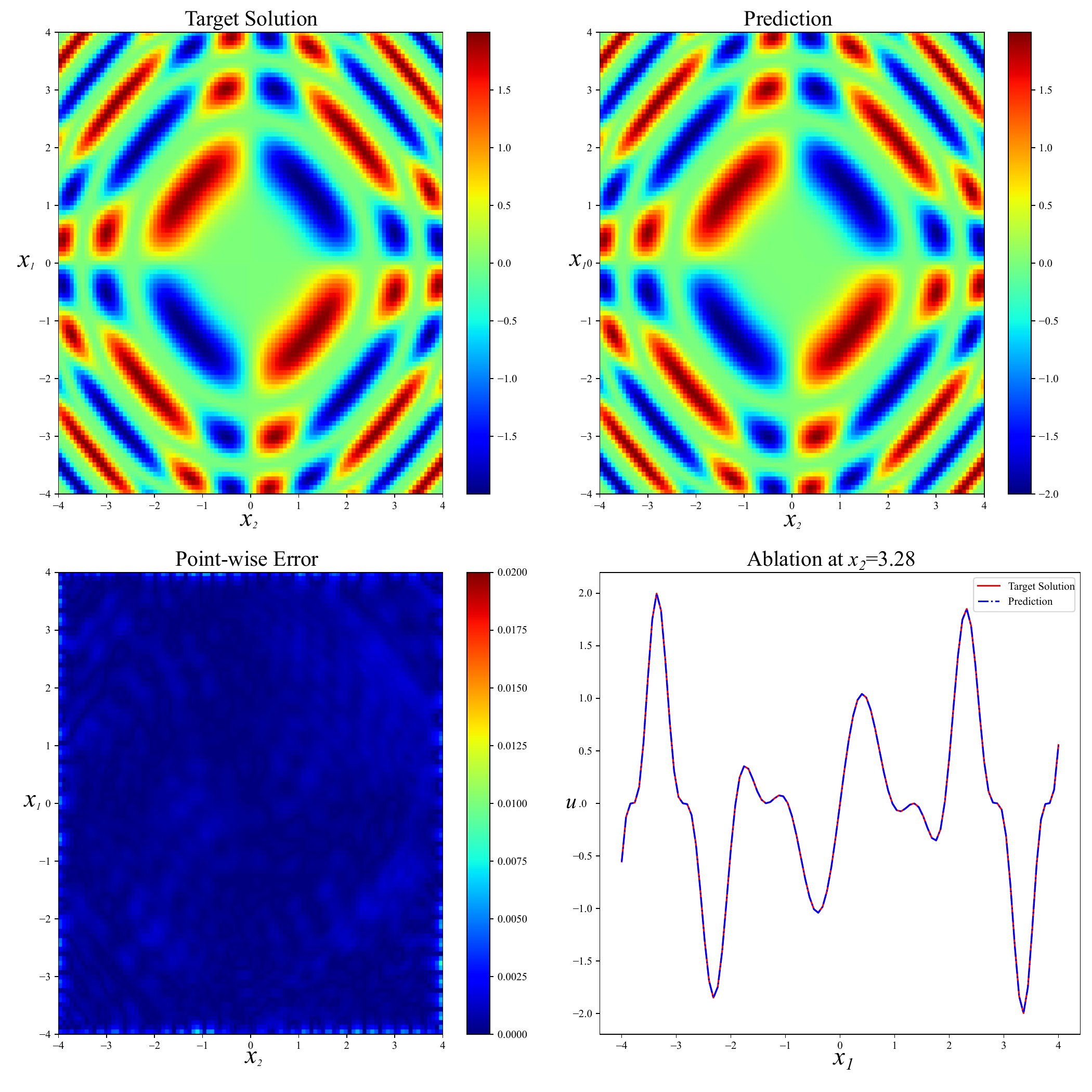}} \\
		(e) AAPINN & (f) Ours
	\end{tabular}
	\caption{\textbf{Visualization of the predicted solutions for the Sine-Gordon equation~\eqref{Sine-Gordon}}. Top left: exact solution; top right: predicted solution; bottom left: point-wise error; bottom right: ablation comparison at $x_2 = 3.28$. }
	\label{SG_Compare}
\end{figure}

Figure~\ref{SG_Compare} presents a visual comparison of predicted solutions for the Sine-Gordon equation across various methods. The proposed AEH method (bottom right) yields remarkably accurate predictions, with visibly lower point-wise error and excellent agreement with the exact solution. Compared to existing baselines, including RAD, SelectNet, CL-Reg, SAPINN, and AAPINN, our method exhibits superior stability and resolution of fine-scale structures. The quantitative accuracy results are summarized in Table~\ref{Table:SG}. Compared with other methods, our strategy achieves the lower relative $ L^2 $ error($ 8.42\times 10^{-4} $). This result validate the capacity of our approach to serve as a reliable and generalizable solver in high-nonlinear, oscillatory PDE scenarios.

\begin{table}[!htb]
	\centering
	\caption{Relative $L^2$ errors of various methods for the Sine-Gordon equation \eqref{Sine-Gordon}.}
	\label{Table:SG}
	\resizebox{\linewidth}{!}{
		\begin{tabular}{ccccccc}
			\toprule
			Method & RAD~\cite{Wu2023} & SelectNet~\cite{Gu2021} & CL-Reg~\cite{Monaco2023} & SAPINN~\cite{McClenny2023} & AAPINN~\cite{Li2024} & Ours \\
			\midrule
			ReL2   & 9.16e-02 & 9.40e-03 & 3.10e-01 & \underline{1.92e-03} & 1.05e-02 & \textbf{8.42e-04} \\
			\bottomrule
		\end{tabular}
	}
\end{table}

\subsection{4D multiscale equation}

Finally, we consider a periodic high-dimensional partial differential equation with multiscale features. Specifically, we study the following equation:
\begin{equation}
\label{4D_equ}
	\begin{aligned}
		\Delta u(\bm{x}) = f(\bm{x}), \quad \bm{x} \in (0,1)^4,
	\end{aligned}
\end{equation}
where the exact solution is given by
\[ u(\bm{x}) = \sin(4\pi x_1)  + \sin(6\pi x_2) + \sin(8\pi x_3) + 0.1\sin(50\pi x_4).\]

This equation poses a significant challenge due to the interplay of high dimensionality and multiscale features, where complex interactions across dimensions and the presence of high-frequency components--especially in  $ x_4 $--make the solution landscape highly intricate. This combination greatly increases the difficulty for numerical and learning-based solvers

\begin{figure}[htbp]
	\centering
	\setlength\fboxrule{0.8pt}
	\setlength\fboxsep{2pt}
	
	\begin{minipage}{0.4\textwidth}
		\centering
		\fbox{\includegraphics[width=\linewidth]{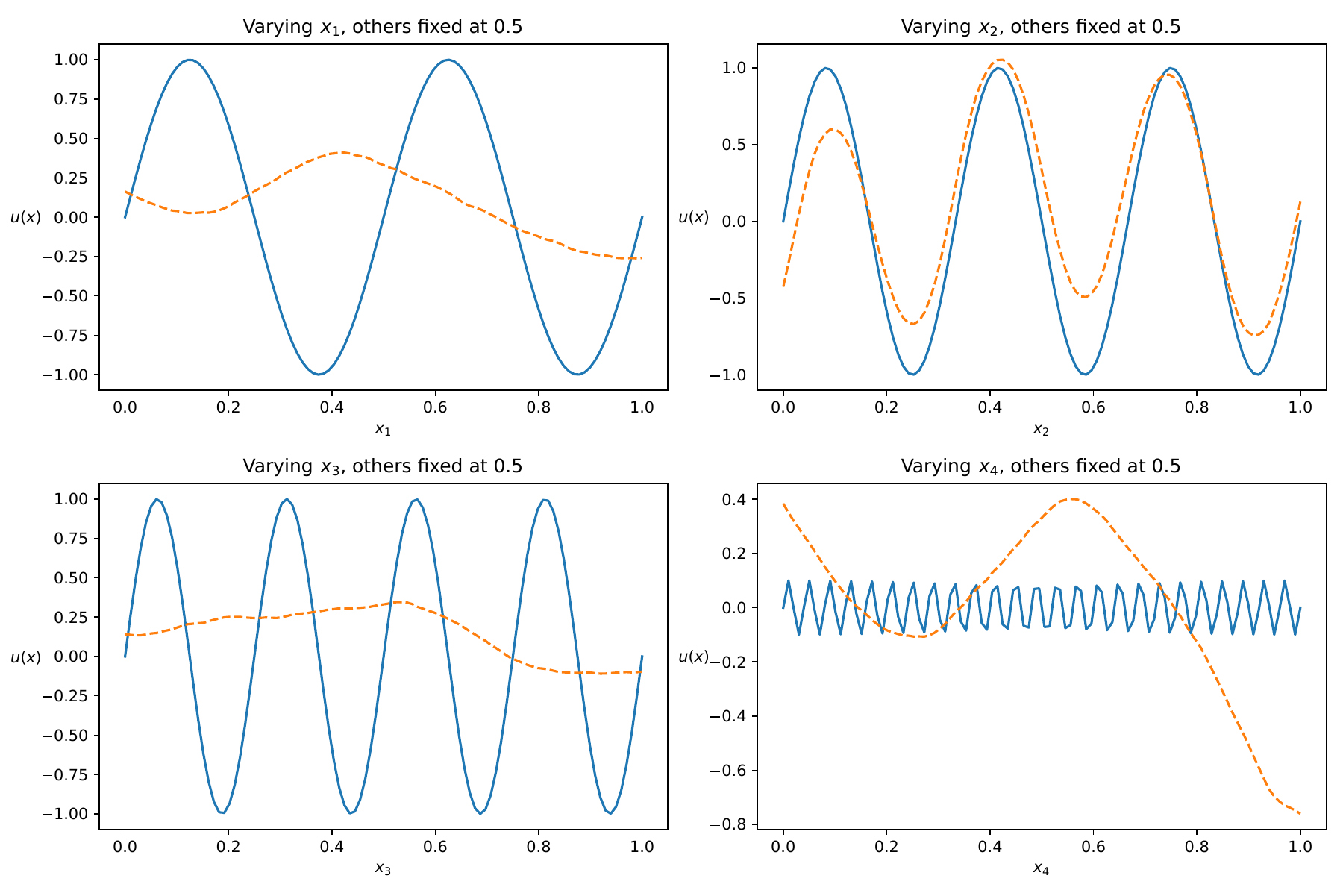}}
		\captionsetup{labelformat=empty}
		\caption*{(a) RAD}
	\end{minipage}
	\hspace{1em}
	\begin{minipage}{0.4\textwidth}
		\centering
		\fbox{\includegraphics[width=\linewidth]{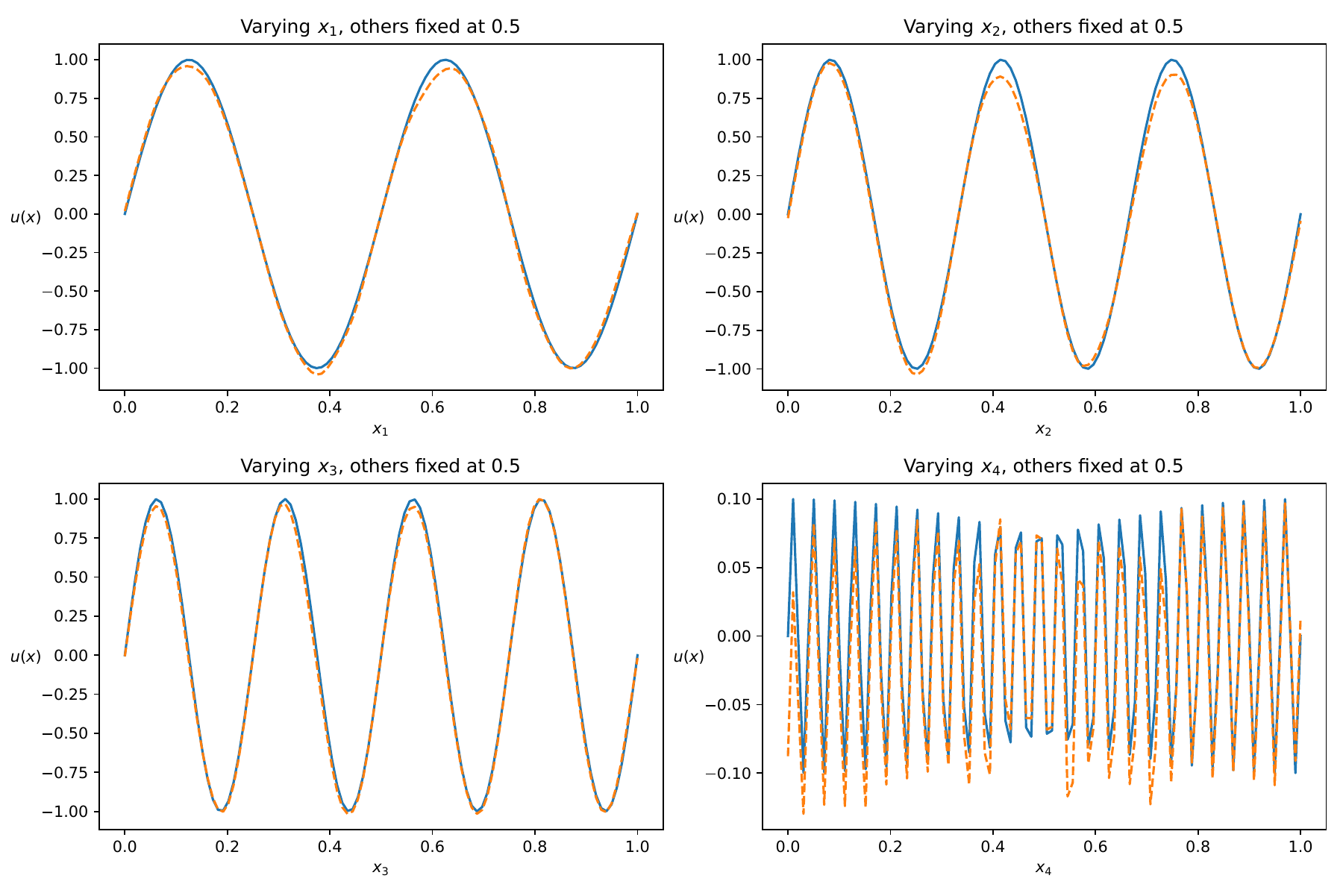}}
		\captionsetup{labelformat=empty}
		\caption*{(b) SelectNet}
	\end{minipage}
	
	\vspace{1em}
	
	\begin{minipage}{0.4\textwidth}
		\centering
		\fbox{\includegraphics[width=\linewidth]{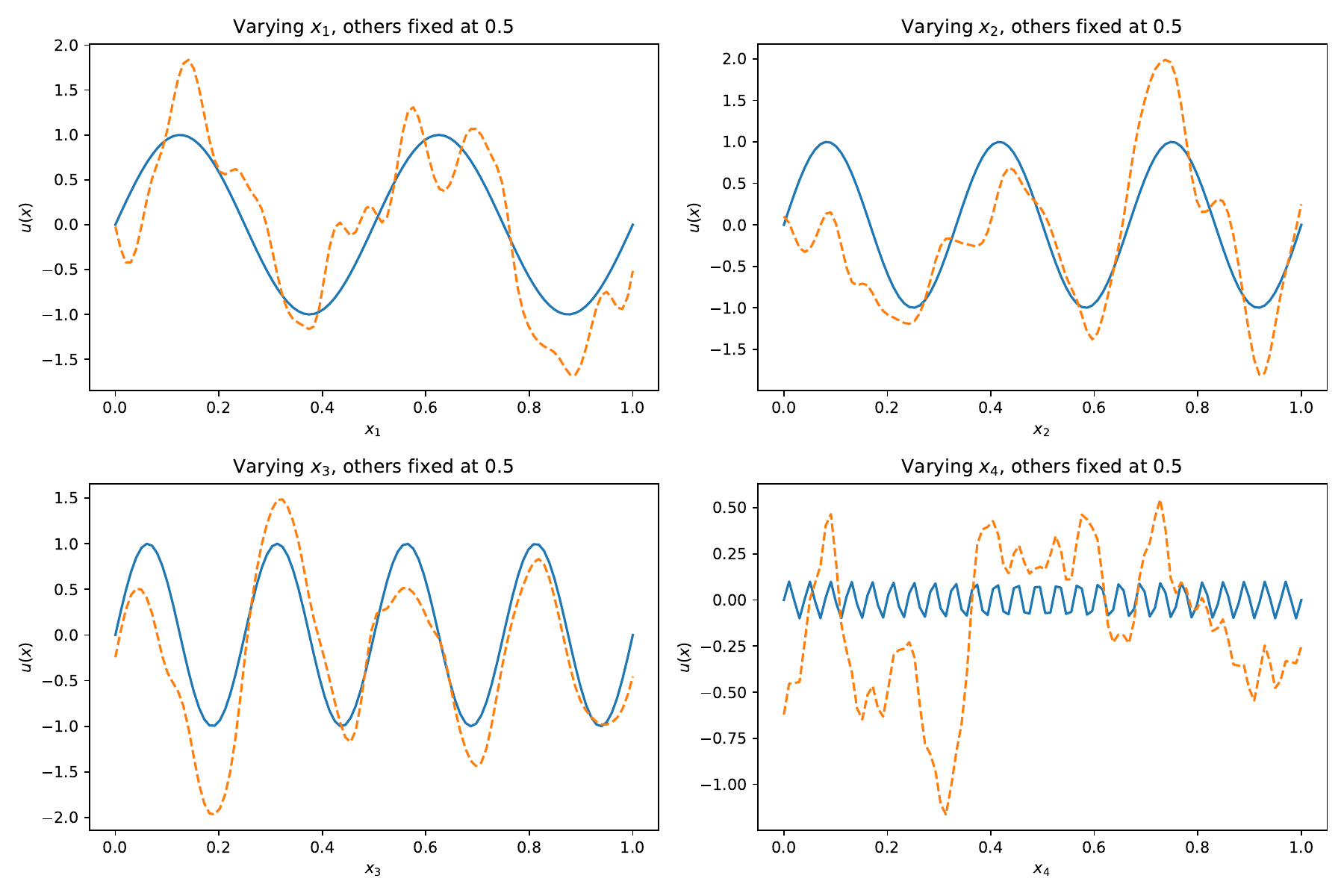}}
		\captionsetup{labelformat=empty}
		\caption*{(c) CL-Reg}
	\end{minipage}
	\hspace{1em}
	\begin{minipage}{0.4\textwidth}
		\centering
		\fbox{\includegraphics[width=\linewidth]{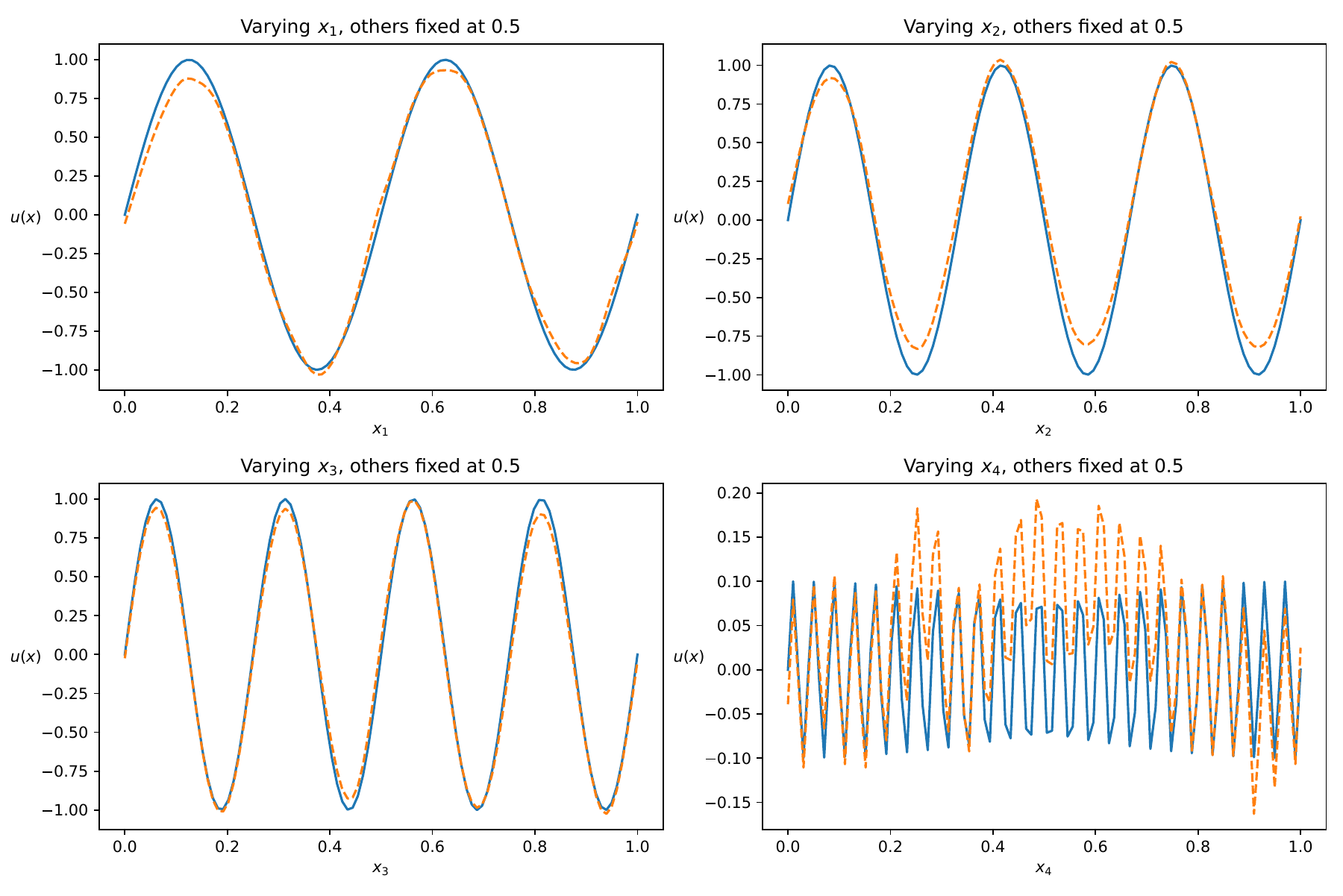}}
		\captionsetup{labelformat=empty}
		\caption*{(d) SAPINN}
	\end{minipage}
	
	\vspace{1em}
	
	\begin{minipage}{0.4\textwidth}
		\centering
		\fbox{\includegraphics[width=\linewidth]{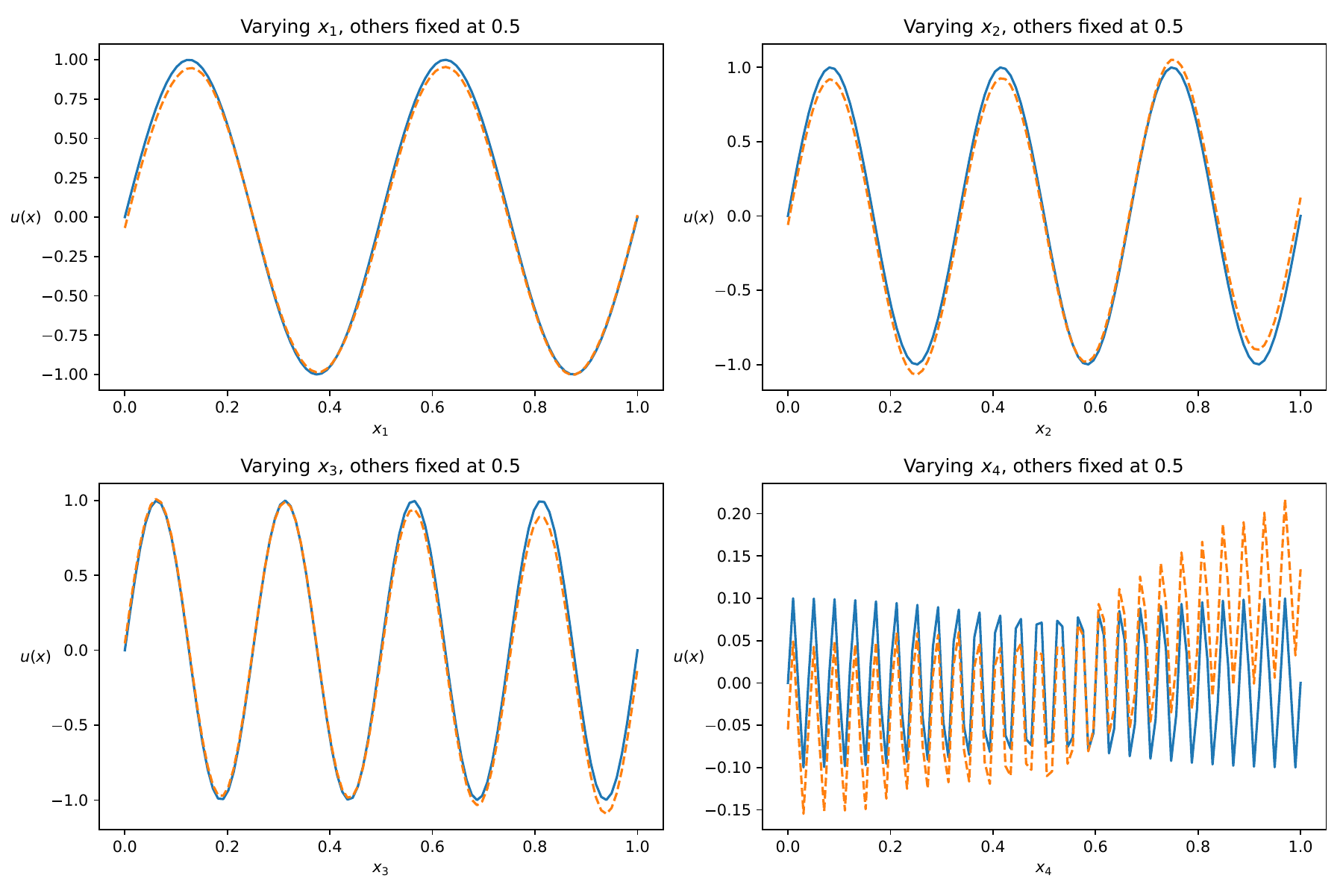}}
		\captionsetup{labelformat=empty}
		\caption*{(e) AAPINN}
	\end{minipage}
	\hspace{1em}
	\begin{minipage}{0.4\textwidth}
		\centering
		\fbox{\includegraphics[width=\linewidth]{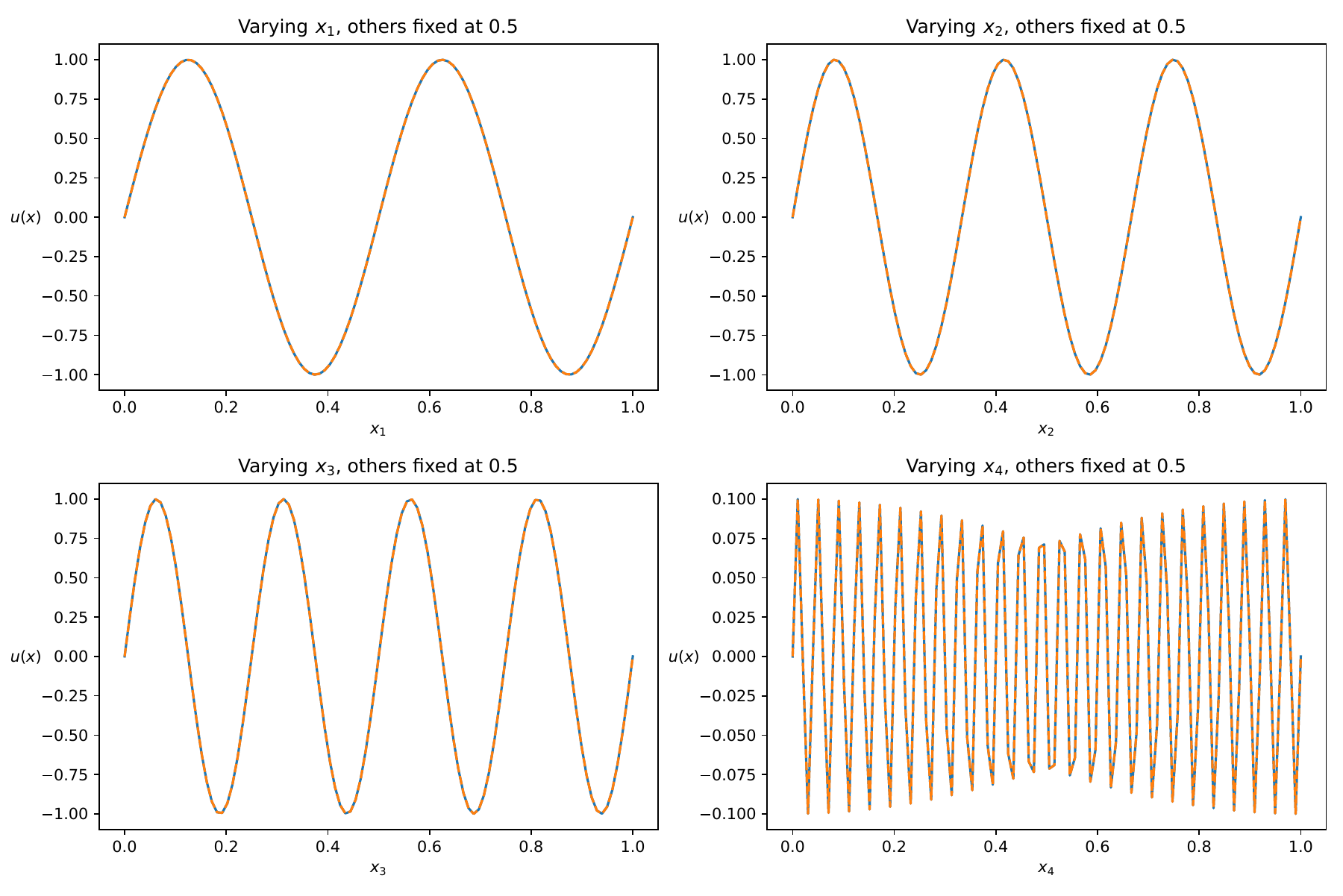}}
		\captionsetup{labelformat=empty}
		\caption*{(f) Ours}
	\end{minipage}
	
	\caption{\textbf{Visualization of the predicted solutions for the 4D multiscale equation~\eqref{4D_equ}}. The blue solid line represents the exact solution, while the orange dashed line denotes the predicted result. To illustrate the solution behavior in each dimension, one variable is varied while the others are fixed at 0.5. Top left: solution profile along $x_1$; top right: along $x_2$; bottom left: along $x_3$; bottom right: along $x_4$. The proposed AEH method accurately captures the complex behavior along the most challenging dimension, $x_4$.}
	\label{4D}
\end{figure}

Figure~\ref{4D} shows the predicted solutions of different methods along each individual dimension, with the remaining three dimensions fixed at 0.5. This visualization evaluates how accurately each method captures the solution's behavior along each coordinate axis. Among these baseline methods, our AEH framework achieves the best performance, especially along the $x_4$ axis, where the high-frequency multiscale feature is most prominent, while other methods struggle to capture these challenging details.

Table~\ref{table:4D} presents the relative $L^2$ errors for all methods. While other baseline methods achieve relative errors on the order of $10^{-1}$ to $10^{-2}$, our method reaches an error as low as \(5.59 \times 10^{-5}\), demonstrating an improvement of three to four orders of magnitude in accuracy. This significant reduction clearly demonstrates the superior accuracy and robustness of our framework in handling high-dimensional, multiscale PDEs.

\begin{table}[!htb]
	\centering
	\caption{Relative $L^2$ errors of various methods for the 4D multiscale equation \eqref{4D_equ}.}
	\label{table:4D}
	\resizebox{\linewidth}{!}{
		\begin{tabular}{ccccccc}
			\toprule
			Method & RAD~\cite{Wu2023} & SelectNet~\cite{Gu2021} & CL-Reg~\cite{Monaco2023} & SAPINN~\cite{McClenny2023} & AAPINN~\cite{Li2024} & Ours \\
			\midrule
			ReL2   & 8.40e-01 & \underline{2.16e-02} & 2.72e-01 & 3.93e-02 & 5.54e-02 & \textbf{5.59e-05} \\
			\bottomrule
		\end{tabular}
	}
\end{table}

\subsection{Impact of alternating mechanism}
Previous experiments show that most baseline methods lack consistent accuracy across different PDEs. In contrast, our method employs a hybrid alternating strategy between a hard prioritization phase (Phase 1) and an easy prioritization phase (Phase 2), leading to more reliable performance. In this subsection, we examine the impact of the alternating mechanism on model performance.

We first conduct an ablation study on three variants: (1) the full alternating model, (2) hard prioritization-only (Phase 1), and (3) easy prioritization-only (Phase 2). As evidenced by Table~\ref{table:ablation_results}, neither single-phase variant achieves consistent dominance, even when one phase outperforms the other on a specific PDE (e.g., Phase 2 excels on Helmholtz, while Phase 1 is competitive for 1D convection-dominated cases). Crucially, the full model not only integrates both phases but also surpasses them individually, reducing errors by 1--3 orders of magnitude across all benchmarks. This indicates that the alternating mechanism is not a simple compromise, but rather a synergistic strategy that achieves superior performance beyond either phase alone.

\begin{table}[htbp]
		\centering
		\caption{Comparative $L^2$ errors: Full mode vs. individual-phase.}
			\begin{tabular}{l c c c }
				\toprule
				Problem 				   	&Full mode 		   & Phase 1  & Phase 2\\
				\midrule
				Heat (steep gradient)  	&\textbf{1.05e-05} &2.90e-02  &\underline{1.26e-03}
				\\
				Helmholtz              	&\textbf{7.61e-05} &3.94e-03  &\underline{4.53e-04}
				\\
				1D convection-dominated	&\textbf{2.79e-06} &\underline{3.90e-06} &5.49e-03
				\\
				Allen-Cahn             	&\textbf{8.12e-05} &3.64e-04 &\underline{2.76e-04}
				\\
				Sine-Gordon            	&\textbf{8.42e-04} &\underline{1.90e-03}  &8.56e-03
				\\	
				4D multiscale             	&\textbf{7.89e-05} &\underline{1.94e-02}  &5.54e-02
				\\	
				\bottomrule
			\end{tabular}
		\label{table:ablation_results}
\end{table}

Second, in our alternating algorithm, each phase is controlled by inner-loop iteration counts $(S_1, S_2)$, with a default setting of $S_1 = 10$ (Phase 1) and $S_2 = 1$ (Phase 2). To examine the influence of iteration counts, we evaluate five configurations: $(1, 1)$, $(1, 10)$, $(5, 1)$, $(10, 1)$, and $(50, 5)$. For fairness, we test on two representative PDEs: the heat conduction equation \eqref{Heat} and the Allen-Cahn equation \eqref{AC}, where the former favors easy prioritization learning, while the latter benefits more from hard prioritization learning.

\begin{table}[ht]
    \centering
    \caption{Influence of inner-loop iteration counts on ReL2 error.}
    \label{table:iteration_impact}
    \begin{tabular}{lcc}
        \toprule
        $(S_1, S_2)$ &Heat (steep gradient) &Allen-Cahn\\
        \midrule
        $(1, 1)$     &2.69e-04  &2.54e-04\\
        $(1, 10)$    & 1.22e-02 &4.11e-03\\
        $(5, 1)$     &1.21e-03  &7.22e-04\\
        $(10, 1)$    & \textbf{1.04e-05}   &\textbf{8.12e-05}\\
        $(50,5)$  	 &\underline{4.18e-05} &\underline{1.08e-04}\\
        \bottomrule
    \end{tabular}
\end{table}

The results in Table~\ref{table:iteration_impact} yield two observations. First, the easy prioritization iterations ($S_2$) should not dominate the hard prioritization iterations ($S_1$); doing so, as in the case of $(1, 10)$, leads to significant error increases. Second, configurations with $S_1 > S_2$, such as $(5, 1)$, $(10, 1)$, and $(50, 5)$, produce similar and stable results. These findings suggest that the hard prioritization phase may play a fundamental role in structuring the solution space, while the easy prioritization phase acts more like a local refinement step. As such, emphasizing hard prioritization iterations ($S_1 > S_2$) can ensure stable convergence and prevent overfitting to simpler patterns.
	
\subsection{Loss curve dynamic}
\label{losscurve}

\begin{figure}[htbp]
	\centering
	\begin{minipage}{0.49\textwidth}
		\centering
		\includegraphics[width=\linewidth]{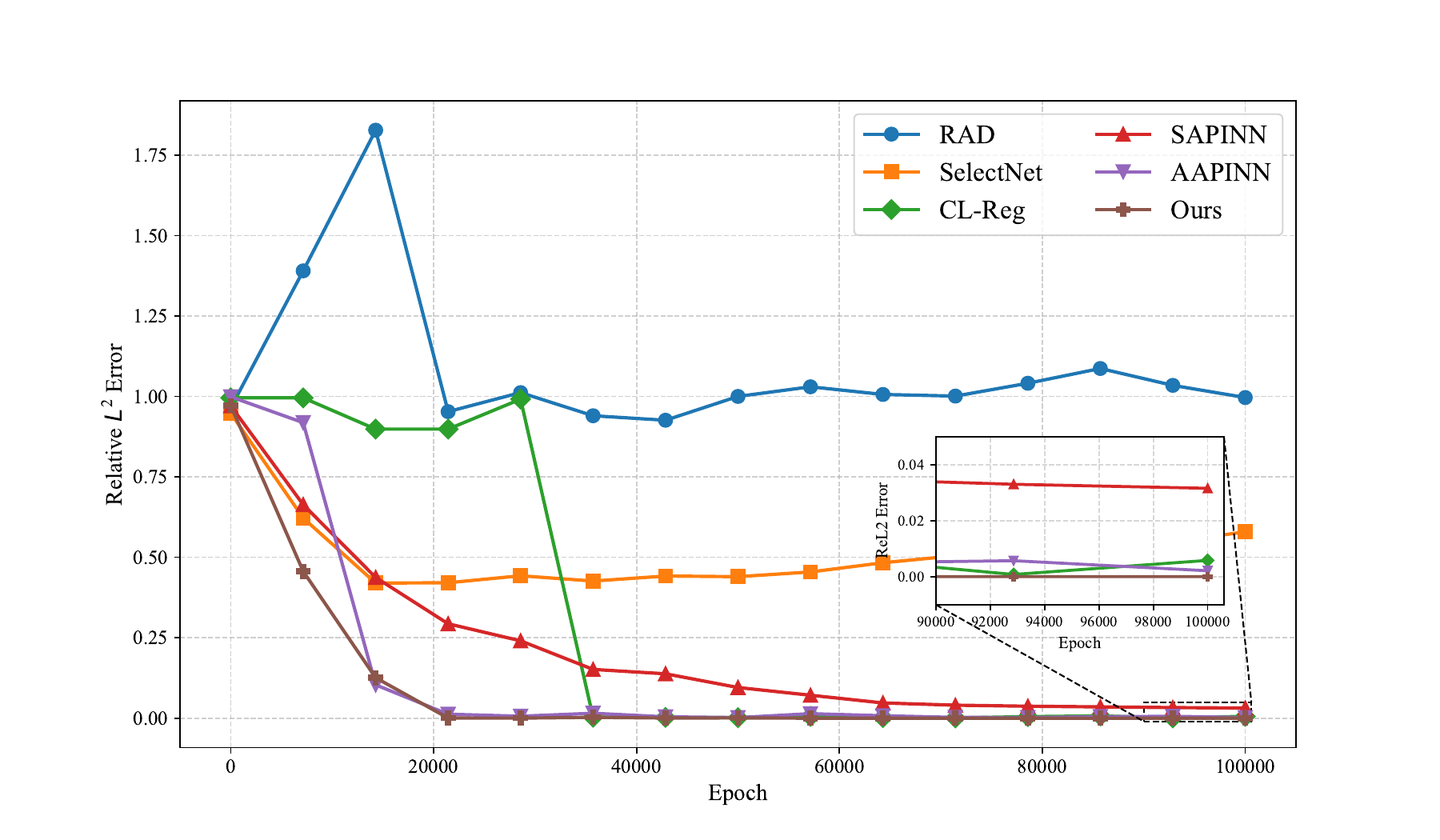}
		\captionsetup{labelformat=empty}
		\caption*{(a) Heat (Steep gradient)}
		\label{fig:heat}
	\end{minipage}
	\hspace{0.17em}
	\begin{minipage}{0.49\textwidth}
		\centering
		\includegraphics[width=\linewidth]{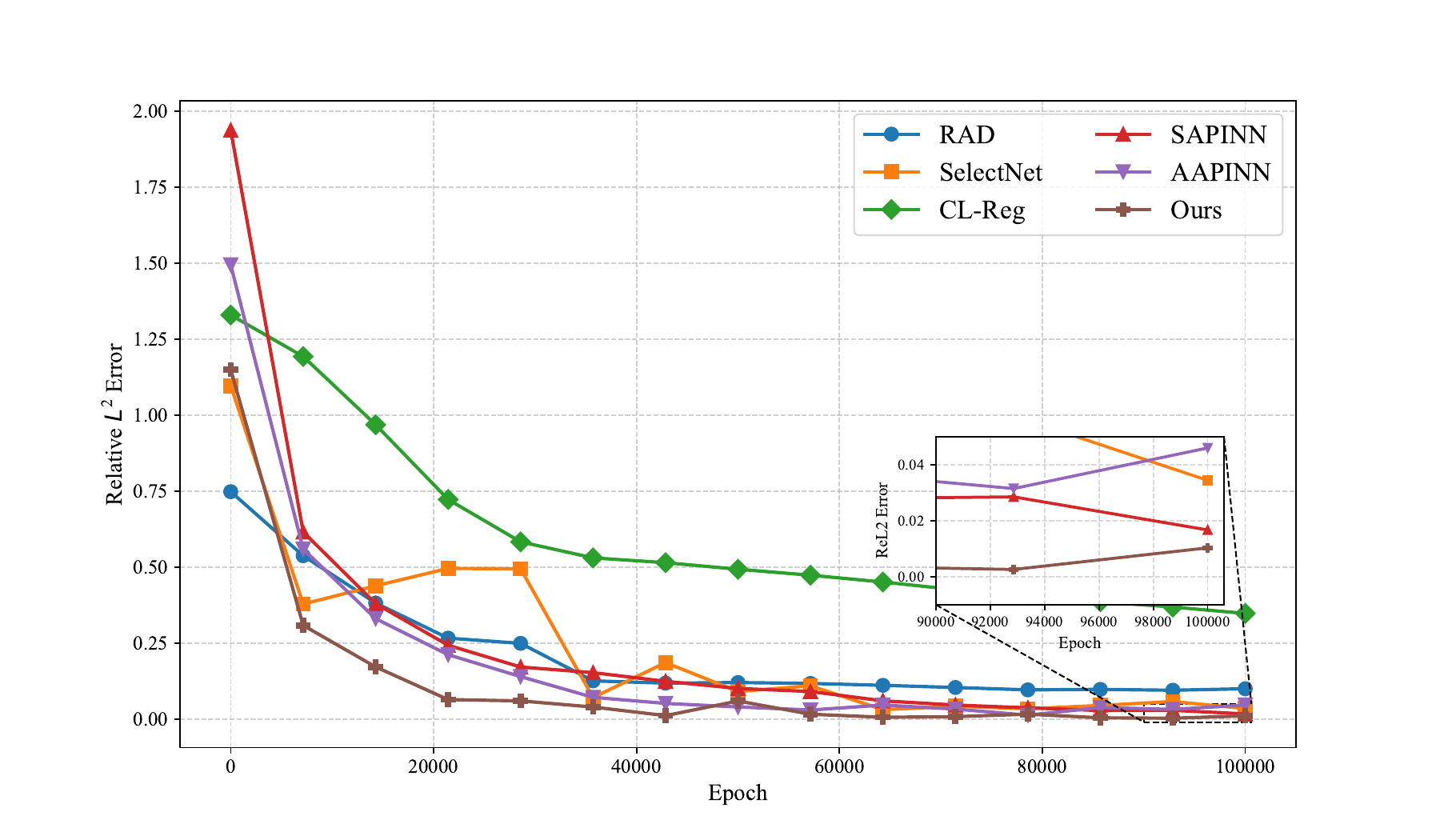}
		\captionsetup{labelformat=empty}
		\caption*{(b) Sine-Gordon}
		\label{fig:AC}
	\end{minipage}
	\caption{Relative $L^2 $ error curves with training epochs.}
	\label{fig:heatmmcompare}
\end{figure}

To further analyze training dynamics, we track the evolution of relative $L^2$ errors during training, as shown in Figure~\ref{fig:heatmmcompare}. Two representative PDEs are considered: (a) the heat conduction equation with steep gradients \eqref{Heat}, and (b) the Sine-Gordon equation \eqref{Sine-Gordon}.

For the heat problem, AAPINN (easy prioritization) achieves a rapid initial drop in error but quickly plateaus, while SAPINN (hard prioritization) improves more gradually yet steadily. In the Sine-Gordon case, AAPINN again converges quickly but stagnates, whereas SelectNet and CL-Reg show significant fluctuations. In contrast, our method achieves faster and more consistent convergence, significantly outperforming all baselines. These observations demonstrate that the alternating strategy not only accelerates early convergence similar to easy prioritization training but also sustains accuracy improvements throughout the entire training process.

\subsection{Computational cost}

\begin{table}[htb]
	\caption{Comparison of training time, model parameter count, and computational load across six models.}
	\label{model_param}
	\centering
		\begin{tabular}{cccc}
			\toprule
			& Time per epoch/s & Params/M & Flops/KFlops \\	
			\midrule
			RAD \cite{Wu2023}
			& 0.0114 & 0.0078 & 7.6
			\\
			SelectNet \cite{Gu2021}
			& 0.0324 & 0.0078 & 7.6
			\\
			CL-Reg \cite{Monaco2023}
			& 0.0139 & 0.0078 & 7.75
			\\
			SAPINN \cite{McClenny2023}
			& 0.4399 & 0.0078 & 7.6
			\\
			AAPINN \cite{Li2024}
			& 0.4718 & 0.0078 & 7.6
			\\
			Ours
			& 0.8799 & 0.0078 & 7.6
			\\
			\bottomrule
	\end{tabular}
\end{table}

Our method incorporates an alternating training mechanism with two phases, which naturally increases training time. As shown in Table~\ref{model_param}, on the Helmholtz equation~\eqref{helm}, our approach needs about twice training time  compared to SAPINN and AAPINN methods. However, the number of parameters and per-pass computational cost (Flops) remain nearly identical across all models, around 7.6 KFlops, which indicates similar architectural complexity and inference-time efficiency.

\section{Conclusion}
Many existing training strategies for PINNs draw inspiration from adaptive algorithms in traditional finite element methods, which often emphasize refining or reweighting samples in regions with large residuals--an approach referred to as hard prioritization. However, deep learning differs fundamentally from classical numerical methods. As a learning-based paradigm, both hard prioritization and easy prioritization strategies can be effective, depending on the problem. In the broader machine learning literature, particularly in computer vision and natural language processing, it is widely recognized that easy prioritization training can outperform hard prioritization training when dealing with highly complex tasks, and vice versa \cite{Graves2017, Jiang2015, WangChen2022, Yang2024}. However, unlike vision or language tasks where difficulty can often be intuitively assessed (e.g., image complexity or sentence length), determining the hardness of PDE solutions is non-trivial in scientific computing, it may depend on hidden factors that are not directly observable from the PDE formulation.

Recent works have introduced easy prioritization strategies to PINNs. Curriculum-based approaches, for example, decompose complex PDEs into a sequence of progressively harder sub-problems \cite{Krishnapriyan2021, Monaco2023}, enabling more stable training. Some studies have proposed down-weighting hard samples to mitigate training instability in PDEs with sharp layers or discontinuities \cite{Li2024, WangXu2024}. Nevertheless, despite these advances, as noted in \cite{Monaco2023}, there is no one rule for PINN. Our analysis confirms this: both hard prioritization and easy prioritization methods exhibit inherent trade-offs and inconsistent performance across different PDEs.

To overcome this, we propose a hybrid alternating training framework that dynamically combines hard and easy prioritization without requiring prior knowledge of sample difficulty. Our approach consistently achieves state-of-the-art accuracy across various challenging PDE benchmarks, reducing relative $L^2$ errors to the order of $10^{-5}$ to $10^{-6}$--improving by several orders of magnitude over existing methods under the same conditions. Importantly, this improvement comes from an effective sample scheduling mechanism without altering the underlying network architecture, making our method widely applicable and easy to integrate.

The current alternating scheme represents one instantiation of hybrid training; exploring other variants may yield further benefits. Additionally, since performance depends on inner-loop hyperparameters, developing adaptive tuning strategies presents a promising avenue for future research.


\end{document}